\documentclass[10pt,twocolumn,letterpaper]{article}

\usepackage[pagenumbers]{cvpr} % To force page numbers, e.g. for an arXiv version

\newtheorem{thm}{Theorem}

\usepackage{multirow}
\usepackage{multicol}
\usepackage{graphicx}

\newcommand{\fx}{f_{x}}
\newcommand{\fxd}{f_{x'}}
\newcommand{\fidic}{FID$_{\text{Inception}}$}
\newcommand{\fiddino}{FID$_{\text{DINOv2}}$}

\newcommand{\kmin}{k$_{\text{min}}$}

\newcommand{\Ntrain}{\text{N}_{train}}
\newcommand{\Nsample}{\text{N}_{sample}}
\newcommand{\Nlearned}{\text{N}_{learned}}

\newcommand{\Xtrain}{{\mathcal{X}}} %Training Dataset 
\newcommand{\Xsynth}{{\mathcal{X}'}} %Snythetic Dataset
\newcommand{\Xlearned}{\mathcal{X}_{\text{learned}}} %Snythetic Dataset

\newcommand{\Pred}{{\mathcal{P}}} %Snythetic Dataset
\newcommand{\Feat}{{\mathcal{F}}} %Snythetic Dataset

\newcommand{\xtraininstance}{{\mathbf{x_t}}} %Real Sample 
\newcommand{\xsynthinstance}{{\mathbf{x'_t}}} %Synthetic Sample 

\definecolor{cvprblue}{rgb}{0.21,0.49,0.74}
\usepackage[pagebackref,breaklinks,colorlinks,allcolors=cvprblue]{hyperref}

\title{Image Generation Diversity Issues and How to Tame Them}
 
\author{Mischa Dombrowski$^1$\qquad Weitong Zhang$^2$\qquad Sarah Cechnicka$^2$ \\
Hadrien Reynaud$^2$ \qquad Bernhard Kainz$^{1,2}$\\
$^1$Friedrich--Alexander--Universit\"at Erlangen--N\"urnberg \qquad
$^2$Imperial College London
\\
{\tt\small mischa.dombrowski@fau.de}}

\begin{document}
\maketitle
\begin{abstract}
Generative methods now produce outputs nearly indistinguishable from real data but often fail to fully capture the data distribution. 
Unlike quality issues, diversity limitations in generative models are hard to detect visually, requiring specific metrics for assessment.
In this paper, we draw attention to the current lack of diversity in generative models and the inability of common metrics to measure this.  
We achieve this by framing diversity as an image retrieval problem, where we measure how many real images can be retrieved using synthetic data as queries. 
This yields the Image Retrieval Score (IRS), an interpretable, hyperparameter-free metric that quantifies the diversity of a generative model's output. 
IRS requires only a subset of synthetic samples and provides a statistical measure of confidence. 
Our experiments indicate that current feature extractors commonly used in generative model assessment are inadequate for evaluating diversity effectively.
Consequently, we perform an extensive search for the best feature extractors to assess diversity.
Evaluation reveals that current diffusion models converge to limited subsets of the real distribution, with no current state-of-the-art models superpassing 77\% of the diversity of the training data.
To address this limitation, we introduce Diversity-Aware Diffusion Models (DiADM), a novel approach that improves diversity of unconditional diffusion models without loss of image quality. 
We do this by disentangling diversity from image quality by using a diversity aware module that uses pseudo-unconditional features as input. 
We provide a Python package offering unified feature extraction and metric computation to further facilitate the evaluation of generative models \href{https://github.com/MischaD/beyondfid}{https://github.com/MischaD/beyondfid}.
\end{abstract}    
\section{Introduction}
\begin{figure}
    \centering
    \includegraphics[width=\linewidth]{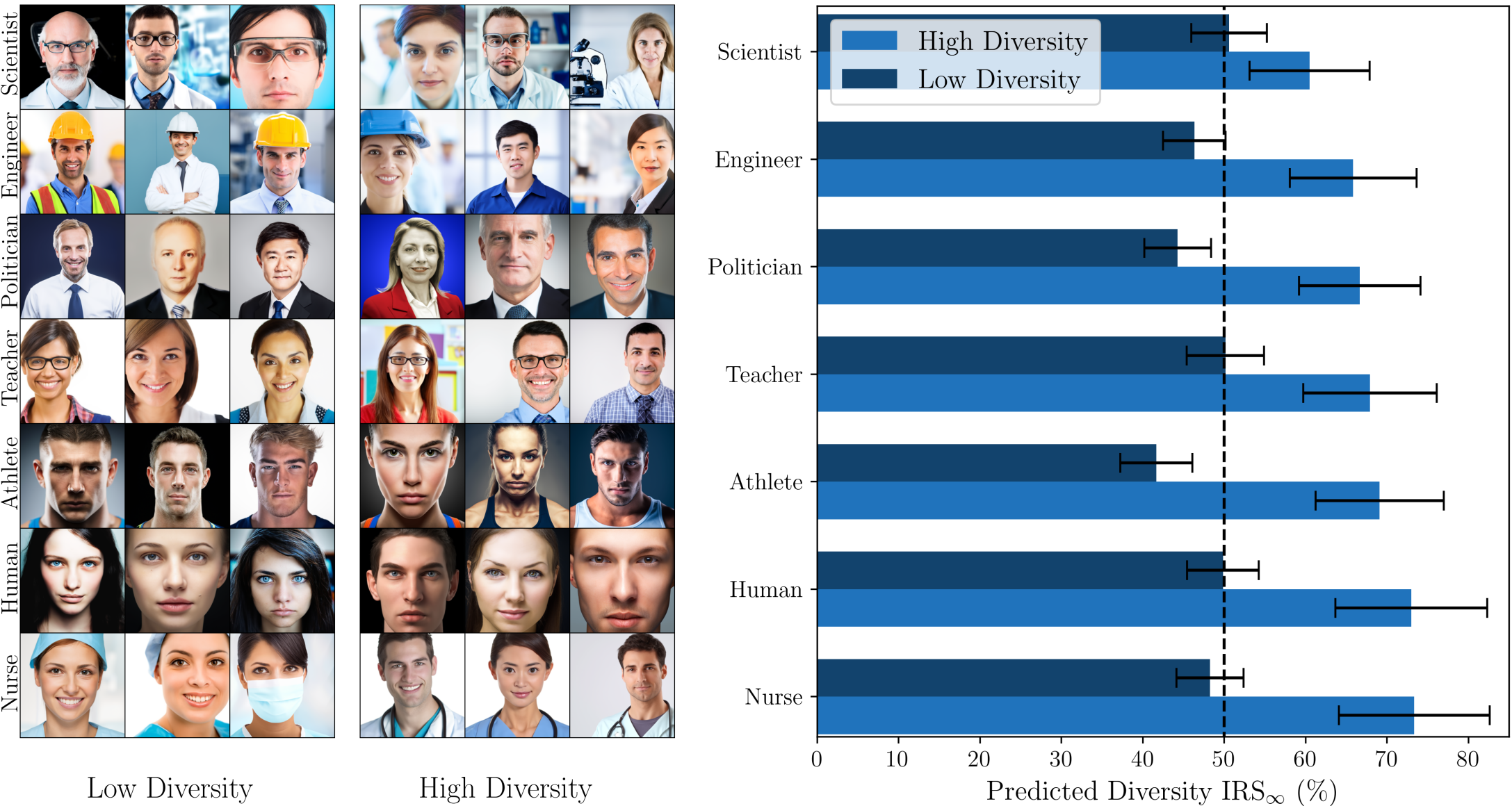}
    \caption{
    Predicted gender diversity - as one possible example for diversity - after sampling pre-trained text-to-image diffusion models infinitely often. 
    Visually it is apparent that most terms are inherently biased toward one gender. 
    Our proposed IRS score predicts that off-the-shelf models will only ever reach 50 percent diversity, which is equivalent to one gender being perfectly represented while the other gender is missing entirely.
    Using diversification strategies increases  diversity. % to more than 50 percent. 
    Details about this can be found in \cref{sec:texttoimagediversity}.
    }
    \label{fig:mainfigdiversity}
\end{figure}
\label{sec:intro}

With advancements in large-scale models and datasets, diffusion models have demonstrated substantial capacity to capture high-dimensional, complex distributions, enabling photo-realistic conditional image generation across diverse guidance types, including image-based~\cite{zhang2024stability}, classifier-based~\cite{dhariwal2021diffusion}, and text-based guidance~\cite{rombach2022high}.
However, diffusion models frequently inherit and amplify data biases due to their reliance on large-scale, naturally biased  datasets~\cite{yang2023diffusion, bansal2022well}, limiting comprehensive distributional coverage in generated samples. For instance, text-to-image diffusion models often exhibit gender bias for certain occupations. 
When prompted with ``a teacher'', models predominantly generate females, with male representations notably underrepresented, as shown in Fig.~\ref{fig:mainfigdiversity}. 
The ``conditional mode collapse''~\cite{aithal2024understanding, miao2024training} also affects image-~\cite{zhang2024stability} and class-conditional~\cite{dhariwal2021diffusion} diffusion models, where the guidance often produces overly uniform outputs, leading to a loss in diversity in the sampled distribution. 
Dataset diversity remains an unresolved issue, even for real datasets \cite{zhao_position_2024}, due to ambiguous definitions and concepts. 
In contrast, synthetic data can leverage its training data as a reference.
To avoid problems like mode drift \cite{alemohammad_self-consuming_2023}, synthetic data should be at least as diverse as its training set. 
Key motivations for ensuring diversity include content creation \cite{ruiz_dreambooth_2022}, addressing privacy and memorization concerns \cite{carlini_extracting_2023}, mitigating the lack of diversity in synthetic datasets \cite{dombrowski_uncovering_2024}, data augmentation \cite{feng_diverse_2023}, promoting algorithmic fairness \cite{friedrich_fair_2023}, and enhancing downstream performance \cite{linguraru_ultrasound_2024}.
To improve diversity in conditional image generation, \cite{liu2024residual} introduces noise perturbations to enhance diversity, though at the expense of image quality and certainty.
\cite{mengsdedit} balances a trade-off between faithfulness and realism in diverse image synthesis, where diversity comes at the cost of fidelity.
These diversity-oriented approaches highlight an inherent compromise: diversity is achieved by sacrificing generation quality, with randomness that lacks interpretability. 
In parallel, classifier-based diversity guidance~\cite{sehwag2022generating} and text-based methods~\cite{ding2021cogview, zhang2023iti} have been developed to reduce bias in text-to-image generation. 
However, these techniques are task-specific and lack generalizability across different types of diffusion models for diversity enhancement.
Thus, a fundamental challenge emerges: \textit{a universal, interpretable metric to evaluate diversity without bias or conditional dependence, and without compromising generation quality.}

To address this, we propose the Image Retrieval Score (IRS), an interpretable metric for quantifying diversity in generative models. 
IRS assesses how effectively synthetic data retrieves real images as queries, providing a hyperparameter-free, statistically grounded evaluation based on minimal samples. 
Our approach is motivated by a thought experiment that illustrates how individual samples reveal model diversity.
\footnote{Consider a die purported to have 1000 sides. If rolled 1000 times but yielding only 10 distinct outcomes, one would naturally question the claim. Similarly, in generative models, each sample serves as a ``roll'', reflecting the model’s diversity in an unconditional manner.}
Using IRS as a core, generic metric for diversity, we quantify and reveal two key findings: (1) existing feature extractors used to evaluate generative models lack the capacity to measure diversity effectively, and (2) current diffusion models converge on limited subsets of the real distribution, with no state-of-the-art models surpassing 77\% of the diversity in training data.

This further motivates the use of IRS as a unbiased guidance to drive a comprehensive search for optimal feature extractors, enhancing diversity evaluation without compromising generation quality based. 
Our selection is based on \cite{stein_exposing_2023} who experimented with on the impact of feature extractors on existing generative metrics. 
Building on these insights, we introduce Diversity-Aware Diffusion Models (DiADM), an approach that expands diversity across the full data distribution in unconditional diffusion models without compromising image quality. 
DiADM accomplishes this by disentangling diversity from image quality through a diversity-aware module utilizing pseudo-unconditional features.
Our contributions can be summarized as follows:
\begin{itemize}
    \item We demonstrate the limitations of widely-used feature extractors for diversity metrics, highlighting their poor performance on real datasets.
    \item We propose an intuitive framework for assessing dataset diversity in unconditional diffusion models.
    \item We identify and quantify the reduction in diversity resulting from current training methodologies.
    \item We provide a comprehensive analysis showing limited image diversity across a wide range of state-of-the-art diffusion models.
    \item We introduce a diversity-aware plug-in module, which utilize pseudo-labels to separate diversity from fidelity, thereby enhancing generative diversity in unconditional diffusion models.
\end{itemize}

\begin{figure}
    \centering
    \includegraphics[width=\linewidth]{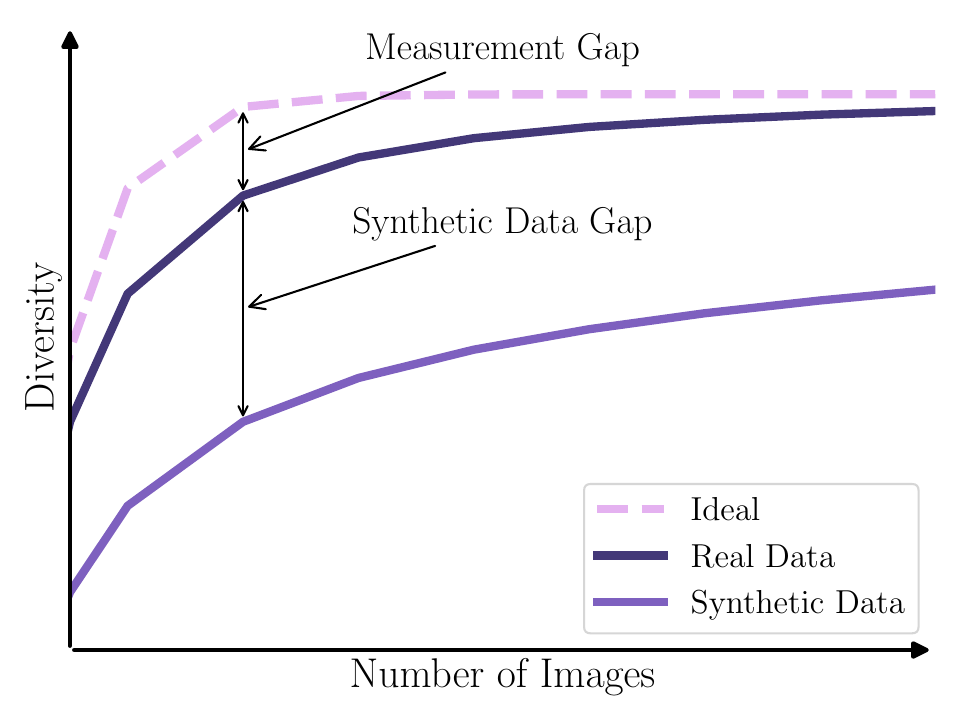}
    \caption{
    We model finding image pairs (image retrieval) as randomly drawing from urnes with replacement (Ideal).
    We observe that all used feature extraction models exhibit performance issues by collapsing in the feature space for real images resulting in a measurement gap affecting currently used metrics such as \fidic, \fiddino, Precision and Recall.   
    Comparing the results of synthetic images to real data shows that datasets generated by generative models show even stronger diversity issues which result in a synthetic data distribution gap. 
    Our proposed metric leverages a real reference dataset to remove this measurement gap. 
    }
    \label{fig:mainfig}
\end{figure}

\section{Related Work}
\label{sec:relatedwork}

\noindent\textbf{Image Generation:}
Diffusion models~\cite{song2020denoising} progressively map complex distributions to a standard Gaussian. 
Conditional diffusion models have demonstrated a strong ability  to generate highly realistic images that align closely with specified conditions, excelling in tasks such as image synthesis~\cite{yang2024improving}, image restoration~\cite{zhang2024unified}, and image reconstruction~\cite{zhang2024stability}. 
Additionally, \cite{dhariwal2021diffusion} enhances diffusion models with an auxiliary classifier, enabling high-fidelity, class-conditional generation, while text-conditional models generate images based on text prompts~\cite{ding2021cogview, zhang2023iti}. 
However, diffusion models often struggle to achieve sufficient diversity in task-agnostic generation, especially when trained on large-scale datasets that carry inherent biases~\cite{yang2023diffusion, bansal2022well}. 
For example, Stable Diffusion, based on latent diffusion models (LDM)~\cite{rombach2022high, radford2021learning}, produces high-quality images aligned with text prompts, yet frequently reflects bias (\emph{e.g.}, age, gender) for certain terms. 
These limitations lead to restricted distributional coverage and biased outputs. 
%Thus, we propose a interpretable diversity metric aimed at enhancing diversity generation in a task-agnostic manner.

\noindent\textbf{Diversity in Generation:}
Research on diversity in diffusion models is limited, with most existing approaches focused on GAN-based models~\cite{chuang2023debiasing, kumari2023multi}. 
Current methods for diversity in diffusion models fall into the primary aims: bias mitigation and diversity enhancement. 
In bias mitigation, \cite{qin2023class} uses a re-weighted loss to address long-tailed distributions, promoting balanced class representation. 
For text-conditional models, \cite{ding2021cogview} attempts to mitigate bias by adding attribute words to prompts, though the added prompts may be ambiguous. 
\cite{miao2024training} extends this to both class- and text-conditional models, using reinforcement learning to match the diversity ratio in reference images but relying on task-specific diversity reward functions.
For diversity enhancement, \cite{liu2024residual} introduces noise perturbations, which increase diversity at the expense of image quality and certainty. 
Similarly, \cite{mengsdedit} balances diversity with realism, where gains in diversity tend to compromise fidelity.
%Our IRS enhances diversity across tasks in diffusion models without sacrificing quality. 
%It serves as a metric for selecting optimal feature extractors and supports the development of a novel Diversity-Aware Diffusion Model (DiADM) that improves distribution coverage within unconditional diffusion models.

\noindent\textbf{Metrics of Diversity:}
Existing metrics in diffusion models remain limited and are unsuitable for accurate and interpretable diversity assessment. 
Traditional image quality metrics like FID~\cite{heusel2017gans} require extensive sampling (\emph{e.g.}, 50k images) to be reliable, making them computationally intensive and challenging to interpret in practical applications.
Precision and recall~\cite{kynkaanniemi_improved_2019} are most established in literature~\cite{rombach_high-resolution_2021,sauer_stylegan-xl_2022,dhariwal_diffusion_2021,peebles_scalable_2023}. 
They add supplementary insights but are hindered by dependency on dataset size, hyperparameters, and interpretability issues. 
Credibility and interpretability of metrics are highly underestimated properties. 
%The absence of these qualities results in a lack of optimization. 
For example, a comparison of three years of advancements in diffusion models, as seen in \cite{dhariwal_diffusion_2021} and \cite{li_autoregressive_2024}, shows an order-of-magnitude improvement in FID but a one percentage point reduction in precision.
Coverage and density~\cite{naeem_reliable_2020} also measure diversity but are dependant on the ratio between train and test set, highly depend on hyperparameters, lack evaluation on real datasets, and are designed to ignore outliers which are important in the context of diversity assesment. 
Vendi Score~\cite{friedman_vendi_2023}, which operates entirely without a real reference set, sacrifices fidelity and is computationally prohibitive for large datasets.
\section{Method}
Let $\xtraininstance$ be image $t$ from a dataset consisting of $\Ntrain$ real images residing in image space $ \Xtrain \in \mathbb{R}^{c \times h \times w}$.
Unconditional generative models aim to learn the distribution $p_{data}(\mathbf{X})$ and sample $\Nsample$ synthetic images from it. 

\subsection{Coupon Collector Problem}
\label{sec:coupon_collector_problem}
The performance of generative models is generally measured by assessing the similarity of the real and the synthetic datasets.  
It is assumed that the training samples represent discrete observations from a continuous unknown distribution. 
We suggest viewing all synthetic images as a composition of real images, which holds true as long as there is no training objective that enforces out-of-distribution generation. 
To assess diversity we argue that every training sample should be the main component of at least one synthetic sample.
To measure this we use the fact that the observed training distribution is discrete. 
A model that has perfectly memorized every training sample would generate all images with equal probability $\frac{1}{\Ntrain}$.
Consequently, we can model sampling synthetic images with an urn experiment. %, which similarly applies to images of models that generalize.
Each synthetic sample consists mainly of one component which is then compounded with other training images.
A real sample is considered \emph{learned} if it serves as the main component for at least one synthetic image. 
Evaluating diversity in this context can be framed as the coupon collector problem~\cite{baum_asymptotic_1965}.
Let $P_n(N_t > 0)$ denote the probability of generating $n$ samples where at least one of the samples corresponds to training image $t$. 
This probability can be computed by considering the compound probability of not sampling this training image which is $\frac{\Ntrain-1}{\Ntrain}$.
Since the same probability applies to each image the expected value of the diversity can be calculated as

\begin{align}
\mathbb{E}[\text{Diversity}] = \sum_{i=1}^{k} \text{P}_n(N_i > 0) \\
= \left( 1 - \left( \frac{\Ntrain-1}{\Ntrain} \right)^n \right).
\label{eq:expecteddiversity}
\end{align}

In this definition, diversity is equal to the number of learned real images. 
This idealized scenario is shown as \emph{ideal} in \cref{fig:mainfig}.
Therefore, assessing diversity comes down to predicting which training image corresponds to which synthetic image. 

\subsection{Image Retrieval}
To assess diversity we first need to define how to decide on what constitutes a main component for each synthetic image. 
For each synthetic image instance $\xsynthinstance$ we aim to identify the real image $\xtraininstance$ that is closest to it according to a measurement $\Pred$ (\emph{e.g.}, Eucledian distance) on the feature of the images extracted by a pretrained feature extractor $\Feat$. 
Then, to assess the diversity we have to compute the size of the dataset of learned images:

\begin{equation}
\Xlearned = \left\{ \xtraininstance \in \Xtrain \mid \exists \xsynthinstance \in \Xsynth : \xtraininstance = \arg \min_{\xtraininstance} \Pred(\xtraininstance, \xsynthinstance) \right\}.
\label{eq:Xlearned}
\end{equation}
We define the cardinality of this set as $\Nlearned$.

We then call the trained image instance $\xtraininstance$ ``learned'' and compute the diversity of the synthetic dataset as: 
\begin{equation}
    \text{IRS}_{\alpha} = \frac{\Nlearned}{\Ntrain} \in \left[0, 1\right]
\end{equation}
We term this quantity as the \textit{image retrieval score (IRS)}.
To emphasize that IRS$_\alpha$ depends on the number of images sampled we add $\alpha := \frac{\Nsample}{\Ntrain}$ as a subscript.  Note that all other generative metrics also depend on $\alpha$ but often ignore this dependency.
The name image retrieval score derives from the fact that computing IRS also works when comparing two disjoined real datasets. In this context the problem can be thought of as an image retrieval task, where a real image serves as the query image and the goal is to find the most similar image in a comparison dataset.  
Intuitively, IRS measures the ratio of images in a training dataset that are retrievable by another dataset.
If the comparison dataset consists of real images we say that an image is ``retrievable''.   
If the comparison dataset consists of synthetic images we say that an image is ``learned''. 

The key advantage of IRS is that we can give a reasonably good estimate of dataset diversity %after just very few samples. 
with only a minimal number of samples.
%If you have retrieved the same image three times after just ten samples you are likely overfitting. 
For example, if the same image appears three times within just ten samples, it is a strong indicator of overfitting.
Consequently, we aim to estimate the diversity after $n << \Ntrain$ samples.
To compute the probability of retrieving exactly $k = \Xlearned$ images we use standard combinatorics. 
This problem can be split into two parts. 
Firstly, we compute the number of possibilities of splitting $n$ into $k$ different subsets. 
Instinctively, we are computing the number of combinations in which different synthetic samples derive from the same retrieved (still unassigned) image $\xtraininstance$ according to equation~\cref{eq:Xlearned}. 
The computation involves using Stirling's number of the second kind, $\text{Stir}(n, k)$, which should be multiplied by the total number of possible assignments for each subset, given by $\frac{\Ntrain!}{(\Ntrain - \Nlearned)!}$.
This will become more clear with an example. 
Say, the first subset consists of three different images $x'_3, x'_{10}, x'_{11}$, all of which are learned from the same real image. 
Then for the first of k subsets, there are $\Ntrain$ different choices. 
For the second subset, we have $\Ntrain-1$ choices and so on. Dividing by all choices leaves us with the probability for generating $k = \Nlearned$ different images:  

\begin{equation}
    \text{P}(k, n, s) = \frac{\text{Stir}(n, k) * s!}{(s - k)! * (s)^n},
    \label{eq:pnlearned}
\end{equation}
with $k = \Nlearned$ and $s = \Ntrain$ and $n = \Nsample$.
For large values, we estimate this by computing the log of \cref{eq:pnlearned} instead. 
\cite{temme_asymptotic_1993} showed that %$\text{Stir}(n,k) = $

\begin{equation}
\text{Stir}(n,k) = \sqrt{\frac{v - 1}{v(1 - G)}} \left( \frac{v - 1}{v - G} \right)^{n-k} \frac{k^n}{n^k} e^{k(1-G)} \binom{n}{k},
\label{eq:symptotic_est_stir}
\end{equation}
where $v = \frac{n}{k}$ and $G \in (0, 1)$ is the unique solution to $G = v e^{G - v}$. 
Asymptotically, this estimate is close, reaching a maximum relative error of roughly $\frac{0.066}{n}$ if n and k are small, \emph{e.g.}, (10, 3)  \cite{temme_asymptotic_1993}.
Computing this in log-space for ImageNet roughly takes 3 seconds and using memorization techniques reduces consecutive computations to 1.3 seconds.

\subsection{Image Retrieval Score} 
Given \cref{eq:symptotic_est_stir} we can now estimate IRS as well as a confidence interval for it.  
We assume that $\Nlearned$ and $\Nsample$ are fixed and we aim to estimate the model diversity. 
To compute the maximum likelihood estimate for IRS we calculate  %$\text{IRS}_{\infty} =$

\begin{equation}
   \text{IRS}_{\infty} = \frac{1}{\Ntrain} \arg\max_{s \in \left[ \Nlearned, \Ntrain \right] }  \text{P}(\Nlearned, \Nsample, s), 
    \label{eq:IRSML}
\end{equation}
 which means we expect $\text{N}_{\text{learned},\infty} = \Ntrain * \text{IRS}_{\infty}$ different images to be sampled eventually. 
 To compute the lower bound we want to minimize the probability of the real diversity being lower but sampling abnormally many different images early. 
 Mathematically, we estimate this as:

\begin{equation} 
\begin{split}
    \text{IRS}_{\infty, L} &= \frac{1}{N_{\text{train}}} \arg\max_{s \in \left[ N_{\text{learned}, \infty}, N_{\text{train}} \right]} \\
    &\quad \sum_{k = N_{\text{learned}}}^{\min(N_{\text{train}}, N_{\text{sample}})} \text{P}(k, N_{\text{sample}}, s) > \alpha_e 
    \label{eq:IRSLower}
\end{split}
\end{equation}

Accordingly, we can compute the estimate for the upper boundary: 

\begin{equation}
\begin{split}
    \text{IRS}_{\infty, U} &= \frac{1}{N_{\text{train}}} \arg\min_{s \in \left[ N_{\text{learned}}, N_{\text{learned}, \infty} \right]} \\
    &\quad \sum_{k = 1}^{s} \text{P}(k, N_{\text{sample}}, s) > \alpha_e
    \label{eq:IRSUpper}
\end{split}
\end{equation}
for a desired probability of error $\alpha_e$. 
%Intuitively, this value computes what N$_{learned,\infty}$ is most likely to have produced $\Nlearned$ different samples after $\Nsample$ sampled images. 
Intuitively, this value represents the most likely estimate of $ N_{\text{learned}, \infty} $ that would produce $ N_{\text{learned}} $ unique samples after drawing $ N_{\text{sample}} $ images. 
We use the infinity symbol to indicate that this estimate represents the potential diversity if we sampled an infinite number of images. 
This value also reflects the model's diversity, representing the percentage of samples that the model can generate at its limit.
In Appx. \ref{sec:model_rejection} we show the dependency of IRS and the confidence intervals on the number of samples showing that at $\alpha = 0.1$, the estimates start to get very confident around the ground truth value.

\subsection{Adjusting for Measurement Gap}
\label{sec:adjustmentstep}
In~\cref{sec:experiments} we will show the measurement gap stemming from feature extractors collapsing to smaller representation spaces. 
This results in the underestimation of diversity for real data which in turn also influences the measured diversity for synthetic data. 
To eliminate the influence of the measurement gap we adjust for the reduced amount of diversity in the feature space of the extractor by normalizing by the measured diversity of real data according to the feature extractor. 
Effectively, we compute $\text{IRS}_{\infty, real}$ on real data using a reference dataset. 
Then we compute $\text{IRS}_{\infty, snth}$ on synthetic data and report the adjusted $\text{IRS}_{\infty, a} = \frac{\text{IRS}_{\infty, snth}}{\text{IRS}_{\infty, real}}$ instead. 
Intuitively, $\text{IRS}_{\infty, a}$ computes the ratio of diversity in the feature space of the snyhtetic images compared to the real images. 
Note that this implies that the adjusted score can be above one for low $\alpha$ values (see \cref{fig:diversity_real}).
We provide additional theoretical analysis in Appx.~\ref{sec:irs over fid} and more quantitative results in Appx.~\ref{sec:other_metrics_hyperparams}, to demonstrate effectiveness in enhancing sensitivity and reliability for measuring diversity in generative models compared to other metrics.

\subsection{Diversity-Aware Diffusion Models}
\label{sec:diversity_awareness_module}
\begin{figure}
    \centering
    \includegraphics[width=\linewidth]{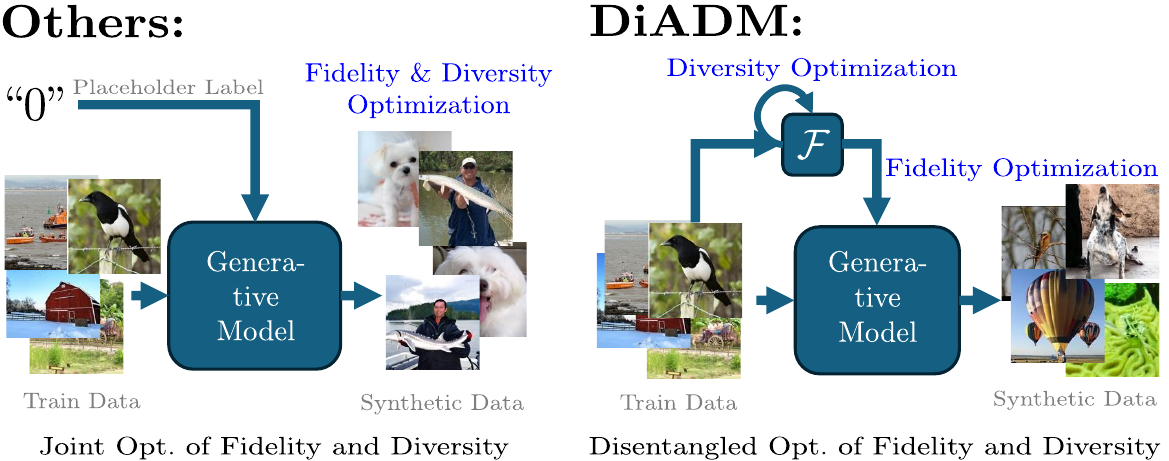}
    \caption{Overview of our proposed DiADM model. Instead of using placeholder labels for unconditional generation, we propose to leverage precomputed features from $\Feat$ instead. That way we can disentangle fidelity (FID) from diversity (IRS).}
    \label{fig:diadm_model_abstract}
\end{figure}
To address diversity issues, we propose DiADMs. 
These models leverage the ability to extract meaningful representations using Inception v3 as feature extractor without relying on labels. 
Rather than employing placeholder labels, as is common in unconditional image generation, we use these pre-computed image features and feed them directly into the model. 
We term this approach ‘pseudo-unconditional’ because, while the architecture resembles a conditional setup, no explicit labels are needed.
In principle, any feature extractor or custom self-supervised model can be used. However, we use the pre-trained ImageNet model, which is widely recognized as an effective feature extractor. 
The core idea is that, with proper training, the model behaves as if each training instance represents its own class, allowing for direct diversification of training instances during sampling time.
Crucially, this approach decouples the diversity of the generated images from their fidelity as shown in ~\cref{fig:diadm_model_abstract}. 
The diffusion model is responsible for producing high-fidelity images, while the diversity module ensures comprehensive coverage of the training distribution’s diversity.
We directly utilize features extracted from $\Feat$ of the training dataset to generate synthetic data, resulting in a synthetic dataset that should maintain diversity if trained properly.
 The backbone model employed is the XS architecture as proposed by \cite{karras_analyzing_2024}, with modifications to adapt the label dimensionality to match the feature dimensionality of $\Feat$. 
Inception is used as the feature extraction model for all datasets to maintain generalizability.

%We also experiment with sampling synthetic features. 
%To maximize the sampling diversity we directly leverage the multivariate normal distribution that is used to compute FID \cite{heusel_gans_2017}. 
%Instead of computing the Wasserstein Distance, we propose generate synthetic features by directly sampling from the distribution learned from the real dataset. 
%Sampling from this distribution is much cheaper than sampling from the generative model itself. 
%Then we can directly reject synthetic features that do not increase the IRS score in feature space without running the costly generative model. 

\section{Experiments}
\label{sec:experiments}
\noindent\textbf{Image Generation Models:} Since diversity is mostly underexplored we aim to compare a series of different state-of-the-art methods against one another. 
\cite{dhariwal_diffusion_2021} introduced classifier guidance and other modifications to Diffusion Models that first lead to competetive performance. 
Unlike all the current methods, they operate in image space. 
\cite{rombach_high-resolution_2021} were the first introduce a model that perform diffusion in latent space. 
\cite{karras_analyzing_2024} use several improved training and design strategies for the U-Net backbone. 
\cite{peebles_scalable_2023} propose to use a Transformer architecture.
\cite{li_autoregressive_2024} use autoregressive prediction an leverage masking for efficient training. 
Our experiments focus on both label-conditional and unconditional generative models, with an emphasis on enhancing diversity in unconditional image generation. 
This focus is motivated by the observation that any form of conditioning ultimately reduces to unconditional generation over the marginal distribution defined by the conditioning.
We use pre-trained models directly and if no unconditional implementation was provided we mimic it by setting the class conditioning to the plaseholder value used for unconditional training \cite{ho_classifier-free_2022}. 
For our experiment with DiADM we train models ourself using a fixed compute budget of 574 A40 GPU hours, which is only one-tenth of the training length in \cite{karras_analyzing_2024}.

\noindent\textbf{Metrics:} We sample each model 50000 times which is a common choice to get a good estimate for FID \cite{karras_style-based_2019,stein_exposing_2023}. 
To mitigate the influence of distribution shift to the test datasets, which is the case for example in ImageNet, we report all metrics on the entire training set. 
To get a refence dataset for the adjusted IRS score, we take away 50000 samples for the reference dataset. 
If the datasets are smaller than 100000 samples we split the dataset in half and average the IRS results for as many runs as we have synthetic images.  

\noindent\textbf{Datasets:} For our primary experiments, we use ImageNet-512~\cite{deng_imagenet_2015}. 
To explore diversity within a more densely populated image space, we also include FFHQ \cite{karras_style-based_2019}, a benchmark dataset focused on facial images, and Chest-Xray14, which comprises medical chest X-ray images \cite{wang_chestx-ray8_2017}.
Additionally, we evaluate CelebV-HQ, a dataset containing face video frames \cite{zhu_celebv-hq_2022}, and Dynamic, a medical video dataset \cite{ouyang_video-based_2020}, treating the data as individual frames as done in \cite{reynaud_echonet-synthetic_2024}.
The spatial resolution is fixed to 512 pixels and videos are additionally limited to the first 60 frames. 

\noindent\textbf{Encoders:} To find the best measurement $\Feat$ we compute the features using a large set of publicly available feature extractors following the selection of \cite{stein_exposing_2023}. 
This includes BYOL \cite{grill_bootstrap_2020}, CLIP \cite{radford_learning_2021}, a ConvNeXT based architecture \cite{liu_convnet_2022}, data2vec \cite{baevski_data2vec_2022}, DINOv2 \cite{oquab_dinov2_2024}, Inception \cite{szegedy_rethinking_2016,heusel_gans_2017,salimans_improved_2016}, MAE \cite{he_masked_2022}, SwAV \cite{caron_unsupervised_2020}. Additionally we add ``Random'' which uses a randomly initialized Inception-v3 \cite{szegedy_rethinking_2016} which has been shown to already extract meaningful featuers \cite{oareilly_pre-trained_2021}.

\subsection{Measurement Gap}
\label{sec:measurement_gap}
\begin{figure}
    \centering
    \includegraphics[width=\linewidth]{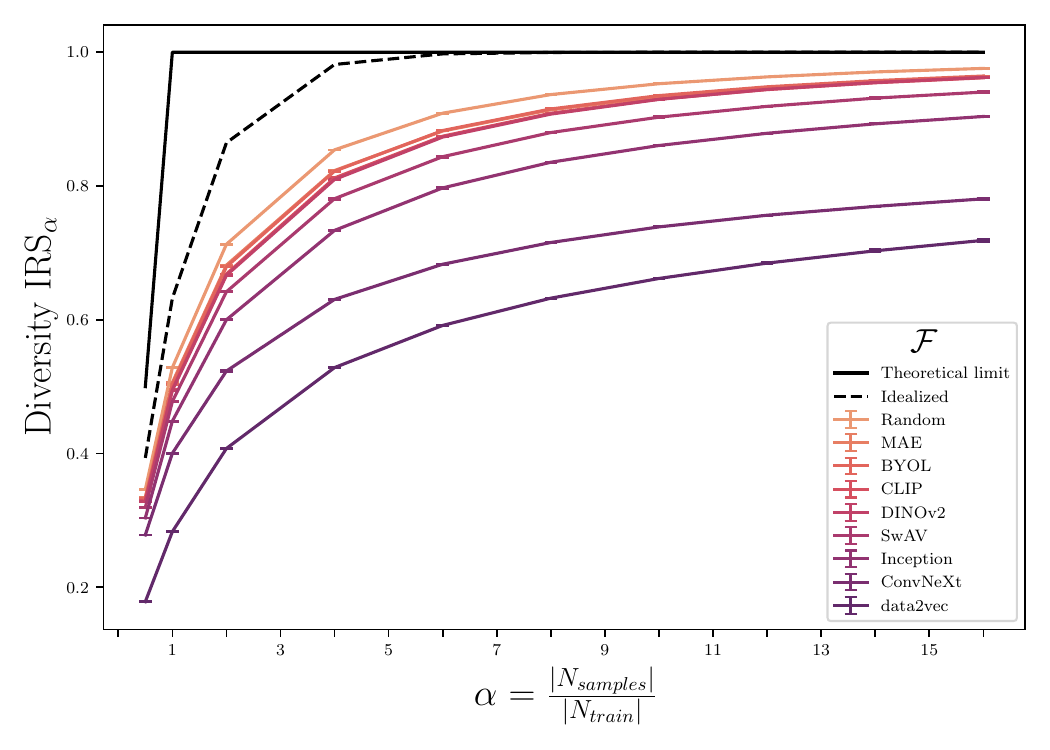}
    \caption{Visualization of the measurement gap across a diverse set of feature extractors by computing the unadjusted diversity of real data. 
    The theoretical limit would be sampling N$_{train}$ images where all of them correspond to a different image in the training dataset. 
    The idealized scenario follows \cref{eq:expecteddiversity}.}
    \label{fig:diversity_real}
\end{figure}

\begin{table*}
    \centering
    \resizebox{\textwidth}{!}{ % Resize the table to fit the text width
    \begin{tabular}{llccccccccc}
    \toprule
    Dataset & N$_{Consensus}$ &BYOL&CLIP&ConvNeXt&data2vec&DINOv2&Inception&MAE&Random&SwAV\\
    \midrule
    ImageNet & 35142 (3.43\%) & 21.72& 89.56& \textbf{95.51}& 2.65& 92.70& 74.20& 87.17& 12.34& 89.39 \\
    FFHQ & 1862 (3.33\%) & 14.18& 93.02& 97.85& 2.42& 92.16& 75.35& 86.79& 21.43& \textbf{98.60} \\
    ChestX-ray14 & 1389 (1.55\%) & 40.32& 45.93& 77.75& 1.94& 83.15& 75.95& 89.70& 33.77& \textbf{96.11} \\
    CelebV-HQ & 5329 (19.47\%) & 31.83& 89.40& 93.71& 6.68& 91.33& 79.94& 94.86& 51.79& \textbf{96.70} \\
    Dynamic & 23 (0.29\%) & 47.83& 34.78& 95.65& 8.70& 78.26& 65.22& 95.65& 65.22& \textbf{100.00} \\
    \bottomrule
    \end{tabular}
    }
    \caption{Measures the agreement of each $\Pred$ with the ensemble if five or more agree on the same retrieved image(\%).}
    \label{tab:agreemen_with_consensus}
\end{table*}

First we need to decide for a measurement $\Feat$ that decides whether two samples correspond to each other. 
We choose to limit ourself to large, pre-trained and publicly available feature extractors due to their universal applicability and established use.
Custom feature extractors are also possible and could be used to increase interpretability even further. 
In Appx. \ref{sec:supp_results_on_echonet} for example we compute IRS using a re-identification model trained to detect memorized samples unveiling diversity issues.
To remain generalizable our goal is to find the most meaningful feature space in terms of image retrieval, that can be generally applied to different datasets. 
We experiment with a vast set of feature extractors $\Feat$, measurements $\Pred$, and datasets.
Importantly we limit the experiments to evaluation on real data. 
In \cref{sec:coupon_collector_problem} we made the assumption, that all samples of the training dataset should be the main component of at least one synthetic image. 
In the context of only comparing real images, this means that every train image should have at least one test image it is more similar to than all other train images.
However, as we show in \cref{fig:barplot-diversity} and \cref{tab:irs} of the supplementary material, all feature extractors are drastically worse than the expected value.
This is a clear indicator for the insufficient representative nature of feature extractors currently used to compute all metrics in the realm of image generation. 
We refer to this observation as \emph{measurement gap}. 
In the supplements we present examples how this measurement gap can be reduced largely if tailord towards certain datasets (\emph{e.g.},~\cref{tab:irs-alpha_results_for_echonetdynamic}). 
However, to remain dataset agnostic we propose an adjustment step that uses the measaured IRS on real data as reference value to make up for this measurement gap. 
Visually this gap results in a lower apparent diversity as shown in~\cref{fig:diversity_real}.
There are two key takeaways from this experiment. 
First, not even real data is diverse according to commonly used features extractors which severly reduces the interpretability of currently used diversity metrics. 
Secondly, it could be worthwhile for special tasks, to train feature extractors specifically to assess diversity as done in \cref{tab:irs-alpha_results_for_echonetdynamic}.

\begin{figure}
    \centering
    \includegraphics[width=\linewidth]{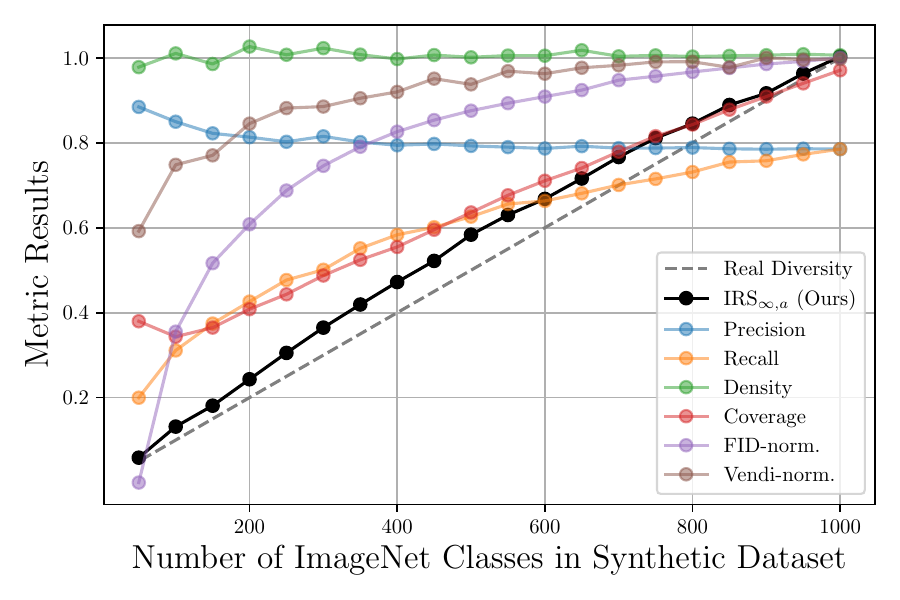}
    \caption{Measuring diversity of datasets by removing classes and computing IRS.
    If only 800 out of 1000 classes are part of the test set say that the diversity is at 80\%.
    By manually removing ImageNet classes we can assess how good commonly used metrics are at measuring diversity compared to ours. To improve visual interpretability we normalize FID and Vendi to be within 0 and 1 with 1 being best.  
    IRS correlates best with the real diversity measured as fraction between number of classes in test and train dataset.}
    \label{fig:metric_interpretable}
\end{figure}

However, thanks to the adjustment step decribed in \cref{sec:adjustmentstep}, for our metric the absolute value of diversity of real data does not matter. 
For synthetic data it just matters how diverse the synthetic data is in comparison to the real data.
This adjustment leads to direct interpretability of the metric as diversity of images. 
In \cref{fig:metric_interpretable} we visualize this interpretability of IRS by balancing and splitting ImageNet into train, reference and test datasets. 
We use the labels to manually reduce the diversity in the test dataset by successively removing classes. 
We use 500 images of each class for train and 100 image of each class for reference and test data.
We can see a clear linear correleation between the number of classes and our proposed IRS score. 
Additionally, it is the only diversity metric that achieves 100\% diversity when all classes are present in the test dataset. 
The slight overestimation of IRS in the mid-diversity-range is likely to be caused by label ambiguity.
Coverage also correlatates well but severely overestimates the diversity when the number of classes is very low. 
It saturates at 0.97 and completely fails for different hyperparmeter settings as we show in \cref{sec:other_metrics_hyperparams}.
Recall, which is the most commonly used metric to assess synthetic diversity, saturates at about 0.8.  
From our point of view, this is one of the key reasons why diversity has not yet established itself as the key metric that should be optimized. 

While IRS works with any kind of feature extractor, we choose to use the one that has the best visual interpretability when it comes to image retrieval to make the results even more meaningful.
Hence, we chooose $\Feat$ based on the meaningfulness of the image retrieval. 
To avoid the necessity of a ground truth image retrieval dataset from the training dataset, we ensemble all feature extractors to compute a consensus on the image retrieval of real images.
Details and qualitative examples can be found in \cref{sec:imageretrieval_agreement_and_consensus}.
Consensus is reached if five or more models agree on the same image pair. 
For every image pair where consensus is reached we compute how often a single model agrees with the ensemble and report the results in \cref{tab:agreemen_with_consensus}.
As expected the percentage of cases where consensus is reached is higher for natural images.
Overall we see that SwAV generally agrees the most with the ensemble. 
Unless mentioned otherwise, this is the feature extractor we use to compure IRS scores.

%DINOv2 has a lower agreement than expected, especially visually, which might be due to the deduplication step applied during training \cite{oquab_dinov2_2024,pizzi_a_2022}.
%All feature extractors are at least eight percentage points away from the expected performance.  
%DINOv2 is best for Imagenet which is in line with \cite{stein_exposing_2023}. 
%Using a randomly initialized model has impressive results throughout all datasets which aligns with \cite{oareilly_pre-trained_2021}. 
%BYOL has impressive results given its self-supervised nature. 
%
%This is what we refer to as \emph{measurement gap}. 
%This is an important indicator for the insufficient representative nature of feature extractors currently used to compute all metrics in the realm of image generation. 
%In the supplements we present examples how this measurement gap can be reduced largely if tailord towards certain datasets. 

\subsection{Diversity Issue in State-of-the-art Diffusion models}
\label{sec:diversity_in_sota_dm}
Recently,~\cite{reynaud_echonet-synthetic_2024} published a synthetic ultrasound video dataset based on the EchoNet dataset~\cite{ouyang_video-based_2020}. 
It is accompanied with a feature extraction model used to privatice the dataset using re-identification. 
Analyzing these features,~\cite{dombrowski_uncovering_2024} were the first to notice a diversity issue. 
Using our proposed IRS score we are now able to quantize it. 
First we try retrieve training samples using real frames from the test dataset. The dataset has 7465 training videos and we can see that 1277 test frames are able to retrieve 1159 distinct train videos. 
Following \cref{eq:IRSML,eq:IRSLower,eq:IRSUpper} this means real data has a diversity of $\text{IRS}_{\infty, real} = 86\% \left[75\%, 100\%\right]$. 
Contrary to that, 1159 different synthetic frames are only able retrieve 692 out of 7465 samples. 
This is equivalent to an $\text{IRS}_{\infty, snth} = 12.3\% \left[11.9\%, 12.9\%\right]$. 
Following the adjustement step introduced in \cref{sec:adjustmentstep} we conclude that the model has only learned $\text{IRS}_{\infty, a} = 14.3\%$ of the data. 
Next, we compare multiple state-of-the-art approaches on conditional and unconditional image generation for ImageNet. We show their diversity accroding to IRS accompanied with other common metrics in~\cref{tab:diversityforgeneratedimages}.
We initially expected that models for lower resolution outperform those with higher resolution. But overall this does not seem to be the case. 
Our experiments on MAR and EDM nicely confirm the feasibility of IRS as metrics as the score increases with the model size. 
Other metrics like Recall and Coverage confirm this observation but the difference seems marginal.
Using IRS we can quantize and interpret the size of the gap between the smallest and largest models. 
The smallest model only reached a diversity of 46\% whereas the largest model reached 75 \%. 
Unlike the relatively small gap of 0.9 points in FID, this indicates a large gap in sampling diversity.
Crucially, following our experiments from \cref{fig:metric_interpretable} our proposed IRS score can be interpreted. 
Even the best model only properly learned 77\% of the data, irrespective of the number of samples generated. 
Note that this is for conditional image generation, so generation specifically asks the model to generate an equal number of images from all classes. 
Generally, the conditional models performed better than the uncondtional model which leads us to believe that the conditioning needs to be improved. 
Due to limited ressources we will continue our experiments by fixing the compute and only use the best performing method in terms of FID and IRS which is EDM.
To keep the training time reasonable we restrict our experiments on the XS variant. 

\begin{table*}[ht]
\label{tab:diversityforgeneratedimages}
\centering
\resizebox{\linewidth}{!}{%
\begin{tabular}{lcccccccc}
\hline
\textbf{Model} & Image Resolution & \textbf{FID} $\downarrow$ & \textbf{Prec.} $\uparrow$ & \textbf{Rec.} $\uparrow$ & \textbf{Dens.} $\uparrow$ & \textbf{Cov.} $\uparrow$ &\textbf{Vendi} $\uparrow$ & \textbf{IRS}$_{\infty,a}$ $\uparrow$ (Ours) \\
\hline
\multicolumn{6}{l}{\textbf{Pixel diffusion}} \\
ADM-256 \cite{dhariwal_diffusion_2021}       & 256  & 6.01 (30.30)& 0.82 (0.57)& 0.62 (0.73)& 1.08 (0.41)& 0.91 (0.40)& 70.94 (36.18)& 0.44 (0.20)\\
\multicolumn{6}{l}{\textbf{Transformer}} \\
DiT-XL/2-256 \cite{peebles_scalable_2023}    & 256  &  22.15 (\textbf{8.72})& 0.94 (0.69)& 0.34 (\textbf{0.76})& 1.58 (0.70)& 0.85 (\textbf{0.84})& 126.96 (\textbf{58.15})& 0.23 (0.33)\\
DiT-XL/2-512 \cite{peebles_scalable_2023}    & 512  &  22.99 (9.54)& \textbf{0.96} (0.70)& 0.27 (0.73)& \textbf{1.90} (0.72)& 0.86 (0.82)& \textbf{128.98} (55.17)& 0.21 (0.34)\\

MAR-B-256 \cite{li_autoregressive_2024}      &256   & 3.79 (10.36)& 0.83 (\textbf{0.72})& 0.67 (0.71)& 1.18 (0.72)& 0.96 (0.75)& 83.03 (55.78)& 0.45 (\textbf{0.38})\\
MAR-L-256 \cite{li_autoregressive_2024}      &256   & 3.30 (10.36)& 0.82 (\textbf{0.72})& 0.71 (0.71)& 1.10 (0.73)& 0.96 (0.75)& 81.80 (55.95)& 0.56 (\textbf{0.38})\\
MAR-H-256 \cite{li_autoregressive_2024}      &256   & 3.11 (10.36)& 0.82 (\textbf{0.72})& \textbf{0.72} (0.71)& 1.07 (\textbf{0.74})& 0.96 (0.76)& 81.37 (55.82)& 0.64 (\textbf{0.38})\\
\multicolumn{6}{l}{\textbf{Latent diffusion, U-Net}} \\
LDM-256 \cite{rombach_high-resolution_2022}  &256   &  26.09 (37.39)& \textbf{0.96} (0.61)& 0.21 (0.68)& 1.80 (0.45)& 0.83 (0.28)& 126.94 (30.83)& 0.16 (0.16)\\

EDM-2-XS-512  \cite{karras_analyzing_2024}    &512  & 3.79 (75.02)& 0.83 (0.42) & 0.65 (0.63)& 1.22 (0.25)& 0.95 (0.13)& 72.41 (26.95)& 0.46 (0.09)\\
EDM-2-S-512  \cite{karras_analyzing_2024}    &512   & 3.33 (122.48)& 0.85 (0.33)& 0.67 (0.42)& 1.26 (0.16)& \textbf{0.97} (0.07)& 80.25 (21.34)& 0.59 (0.04)\\
EDM-2-M-512  \cite{karras_analyzing_2024}    &512   & 3.30 (107.45)& 0.85 (0.36)& 0.69 (0.61)& 1.24 (0.19)& \textbf{0.97} (0.09)& 82.99 (22.19)& 0.65 (0.06)\\
EDM-2-L-512  \cite{karras_analyzing_2024}    &512   & 2.90 (118.87)& 0.84 (0.23)& 0.70 (0.51)& 1.22 (0.11)& \textbf{0.97} (0.06)& 82.10 (22.89)& 0.71 (0.03)\\
EDM-2-XL-512  \cite{karras_analyzing_2024}    &512  & 2.92 (141.74)& 0.84 (0.25)& 0.71 (0.45)& 1.21 (0.12)& \textbf{0.97} (0.06)& 83.23 (20.04)& \textbf{0.77} (0.03)\\
EDM-2-XXL-512  \cite{karras_analyzing_2024}    &512 & \textbf{2.87} (124.29)& 0.84 (0.33)& 0.71 (0.60)& 1.22 (0.17)& \textbf{0.97} (0.07)& 82.45 (21.52)& 0.75 (0.05)\\

\hline
\end{tabular}%
}
\caption{Image generation of ImageNet across different generative models. 
Numbers in brackets indicate results when sampling the model without conditioning. 
Inception is used as feature extractor for all metrics except for IRS where we use SwAV. 
Precision, recall, density, coverage and Vendi score had to be computed on a subset of the training data of 50000 and 10000 for Vendi score. 
Pretrained conditional models were used as unconditional models by setting conditional guidance to 0.}
\end{table*}

\subsection{Improving Diversity Using Pseudo Labels}
Now that we have established EDM as the best method for unconditional image generation and at the same time have shown that this method still lacks diversity, we continue with EDM-2 as baseline for unconditional image generation and compare the trained models with a fixed compute budget. 
Quantitative results are shown in \cref{tab:generative_results_other}. 
The results show that the leveraging pseudo labels improves the sampling diversity compared to EDM-2 in all cases.
In three cases the diversity is even better than that of the real reference dataset proofing that the pseudo conditioning works well and we can specifically query the model to generate diverse data. 
It also improves FID scores compared to the unconditional trainings.

\begin{table}[ht]
\resizebox{\linewidth}{!}{%
\begin{tabular}{lcccc}
\toprule
        & \multicolumn{2}{c}{FID $\downarrow$ } & \multicolumn{2}{c}{IRS$_{\infty,\text{a}}$  $\uparrow$ } \\
        \cmidrule(r){2-3}\cmidrule(r){4-5}
        & EDM   \cite{karras_analyzing_2024}  & DiADM (Ours) & EDM \cite{karras_analyzing_2024} & DiADM (Ours) \\
%Conditioning & Uncond. & Pseudo-Cond. & Uncon. & Pseudo-Cond.  \\
%        \midrule
ImageNet-512 & 51.59 & \textbf{22.28}  &0.09 & \textbf{0.15}\\
FFHQ & 40.92         & \textbf{6.24}   &0.23 & \textbf{1.51}\\
ChestX-ray14 & 24.29 & \textbf{6.76}   &0.19 & \textbf{1.08}\\
CelebV-HQ & 68.41    & \textbf{13.64}  &0.18 & \textbf{0.69}\\
Dynamic & 13.82      & \textbf{5.56}   &0.60 & \textbf{1.04}\\
\bottomrule
\end{tabular}%
}
\caption{FID and IRS results with and without our proposed diversity awareness module}
\label{tab:generative_results_other}
\end{table}

%%%Only if enough time and paper space
\subsection{Text-to-image example} 
\label{sec:texttoimagediversity}
Extending IRS to text-conditional image generation is straightforward. 
The evaluation of diversity depends on the chosen perspective, which is defined by the reference datasets. 
By changing the reference dataset, we can address various diversity-related issues, such as fairness. 
One notable example is the gender bias observed in text-to-image models. 
For our experiments, we use Deepfloyd (IF-I-XL-v1.0)\footnote{https://huggingface.co/DeepFloyd/IF-I-XL-v1.0}. 
To assess gender diversity, we create a reference dataset with a balanced gender distribution for specific job roles by prompting the model to generate 100 images each of males and females. 
Half of these images are reserved as a balanced test dataset. 
We then prompt the model without specifying gender. 
The results are presented in \cref{tab:diversityforgeneratedimages}. 
All groups, including the general term \emph{human}, show gender bias, with the latter generating only female images. 
IRS estimates indicate that in all cases, the generated diversity reaches only about 50\% of that of the balanced reference dataset.

\subsection{Limitations}
The stochasticity involved for a low number of samples means that minor IRS differences are not meaningful for a low number of samples. 
However, this is also reflected in the high values of uncertainty.
Additionally, we believe that finding a better feature extractor that maximizes measured diversity (unadjusted IRS) on real datasets should be the focus. 
Experiments on improving diversity are limited to using real features for generation and restrictive computational budget. 
Furthermore, increasing the conditioning requires similar experiments analyzing memorization issues as text-conditioning~\cite{carlini_extracting_2023}.
\section{Conclusion and Outlook}
In this paper, we reveal that all current methods for evaluating diversity  and the feature extractors they are based on are inadequate. 
Using our proposed IRS metric, we demonstrate that these shortcomings results in all state-of-the-art image generation methods facing challenges with diversity. 
To address this issue for unconditional diffusion models, we propose DiADMs, which enhance the performance of current methods by separating diversity from fidelity. 
In future work, we aim to extend our efforts to improve diversity in conditional image generation as well.

\noindent\textbf{Acknowledgements:} The authors gratefully acknowledge the scientific support and HPC resources provided by the Erlangen National High Performance Computing Center (NHR@FAU) of the Friedrich-Alexander-Universität Erlangen-Nürnberg (FAU) under the NHR projects b143dc and b180dc. NHR funding is provided by federal and Bavarian state authorities. NHR@FAU hardware is partially funded by the German Research Foundation (DFG) – 440719683. Support was also received from the ERC - project MIA-NORMAL 101083647 and DFG KA 5801/2-1, INST 90/1351-1. This work was supported by the UKRI Centre for Doctoral Training in Artificial Intelligence for Healthcare (EP/S023283/1) and Ultromics Ltd.
{
    \small
    \bibliographystyle{ieeenat_fullname}
    \bibliography{main,references}
}

\clearpage
\maketitlesupplementary

\section{Model Rejection Based on IRS}
\label{sec:model_rejection}
\begin{figure}
    \centering
    \includegraphics[width=\linewidth]{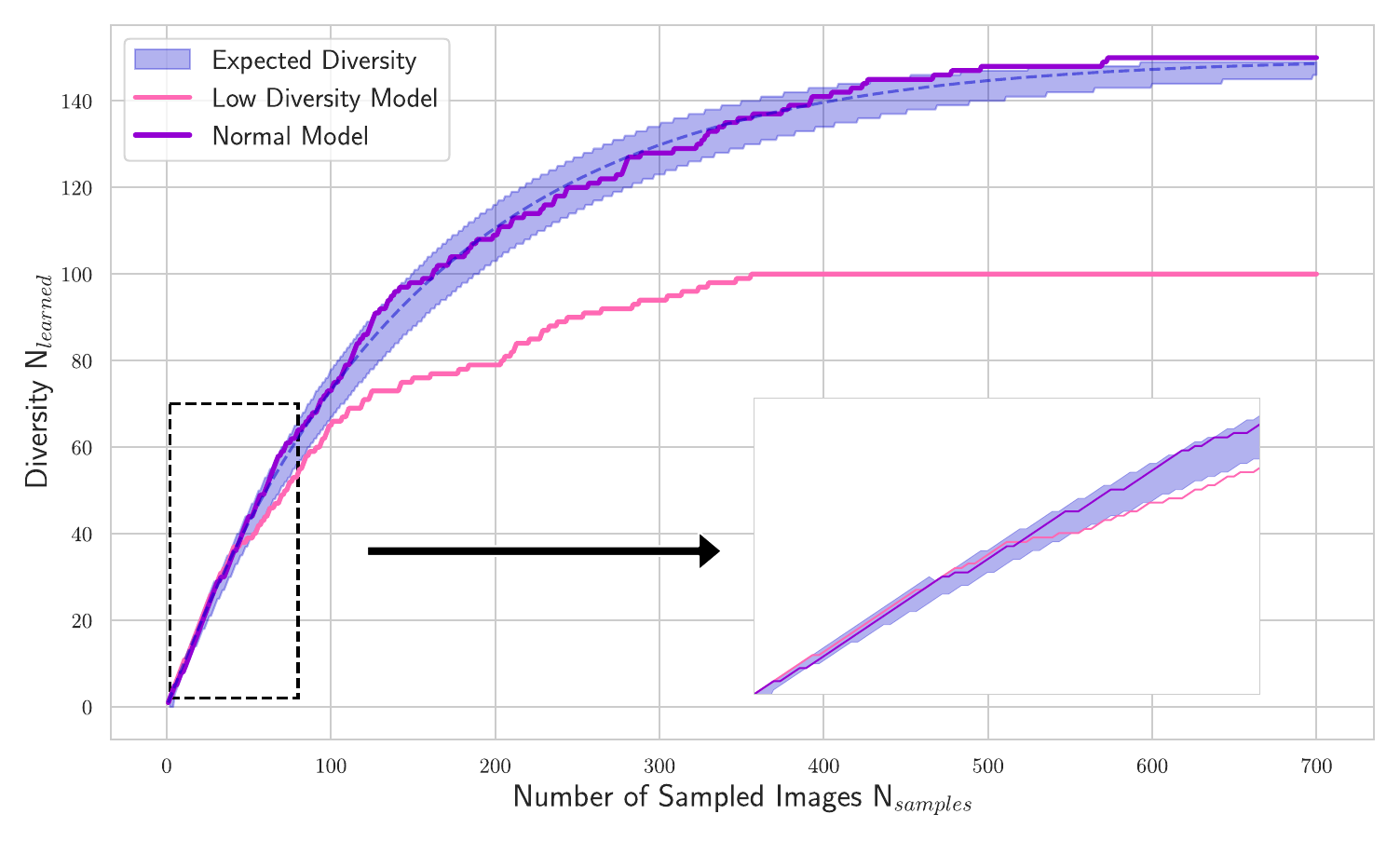}
    \caption{Detecting low diversity models. %Knowing training dataset statistics we can estimate the diversity of the model while sampling. 
    By leveraging the statistical properties of the training dataset, we can assess the diversity of the model during the sampling process. 
    A low diversity model will saturate earlier and can be rejected quickly. In this example, we would reject the low diversity model after only generating 60 samples.}
    \label{fig:detectinglowdiversity}
\end{figure}

Building on the methodology introduced in the main paper, we explore how IRS can provide additional insights into model diversity during training. 
IRS, which requires a minimal number of samples for computation, proves to be particularly useful for online evaluation methods such as rejecting early model checkpoints that lack sufficient diversity.
This appendix analyzes IRS-based rejection through experiments, providing practical examples and additional interpretation of its efficacy.
By defining a target diversity percentage $\text{IRS}_d$ , we can leverage IRS to monitor progress towards achieving this goal. 
The core question IRS addresses is: 'How 
many images learnt from different training samples~\kmin will a model with this level of diversity generate at minimum?'
To calculate this, we take~\cref{eq:IRSLower,eq:IRSUpper}, and look for the minimum expected number of different learned images \kmin:

\begin{equation}
 \text{k}_{\text{min}} = \arg \max_{k} \sum_k \text{P}(k, \Nsample, \text{IRS}_d * \Ntrain) < \alpha_e
\label{eq:kmin}
\end{equation}
\cref{fig:detectinglowdiversity} illustrates this using a toy example with 800 distinct training values. 
To get ground truth values for diversity, we simulate generation as random sampling from a set of integers. 
%We measure the number of different learned samples $k = \Nlearned$ and compute~\kmin.
As shown in \cref{fig:detectinglowdiversity}, a toy example with 800 training images demonstrates how low-diversity models quickly converge to a high number of duplicates. 
By sampling as few as 60 images, we can confidently reject models failing to meet the diversity threshold. 
This highlights IRS as a reliable tool for early-stage model evaluation.

\begin{figure}
    \centering
    \includegraphics[width=\linewidth]{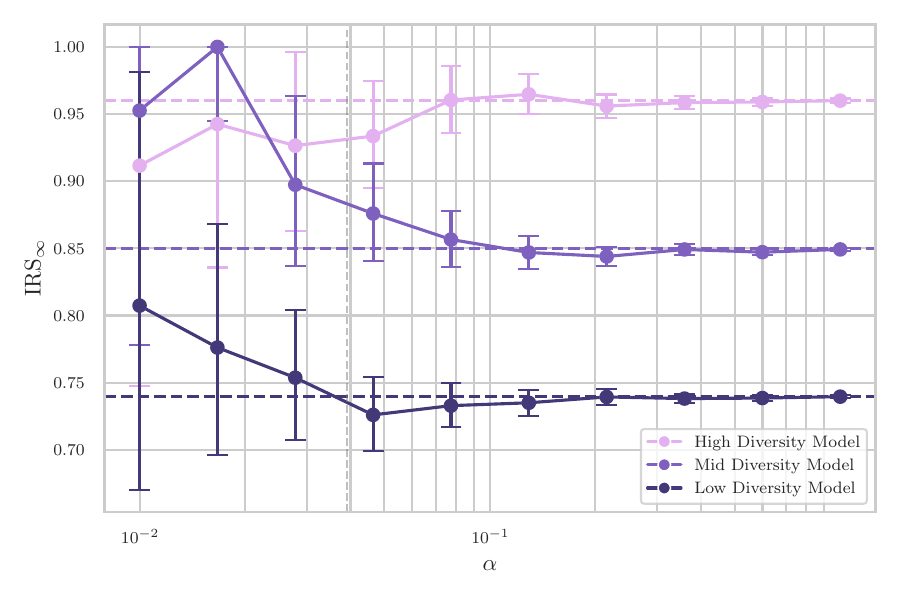}
    \caption{IRS$_\infty$, indicated by dots, and confidence intervals IRS$_{\infty,L}$ and IRS$_{\infty,U}$ for three different simulated ImageNet models. 
    By increasing the number of observations ($\alpha$), we also increase the confidence of the IRS$_\infty$ predictions. 
    The black dashed line indicates the ground truth and the vertical line the value for 50k samples.}
    \label{fig:irs_infinity}
\end{figure}

We additionally investigate the efficacy of IRS by examining its accuracy according to \cref{eq:IRSML,eq:IRSUpper,eq:IRSLower}. 
Figure \ref{fig:irs_infinity} illustrates how predicted IRS values vary with model diversity.
The initial estimate with $\alpha = 0.01$, which evaluates the diversity of $1/100 * \Ntrain$ samples, is reasonably close to the true value, but its confidence level suggests this may be due to chance.
At $\alpha = 0.1$ the estimates become increasingly confident around the ground truth value.

\begin{figure*}
    \centering
    \includegraphics[width=\linewidth]{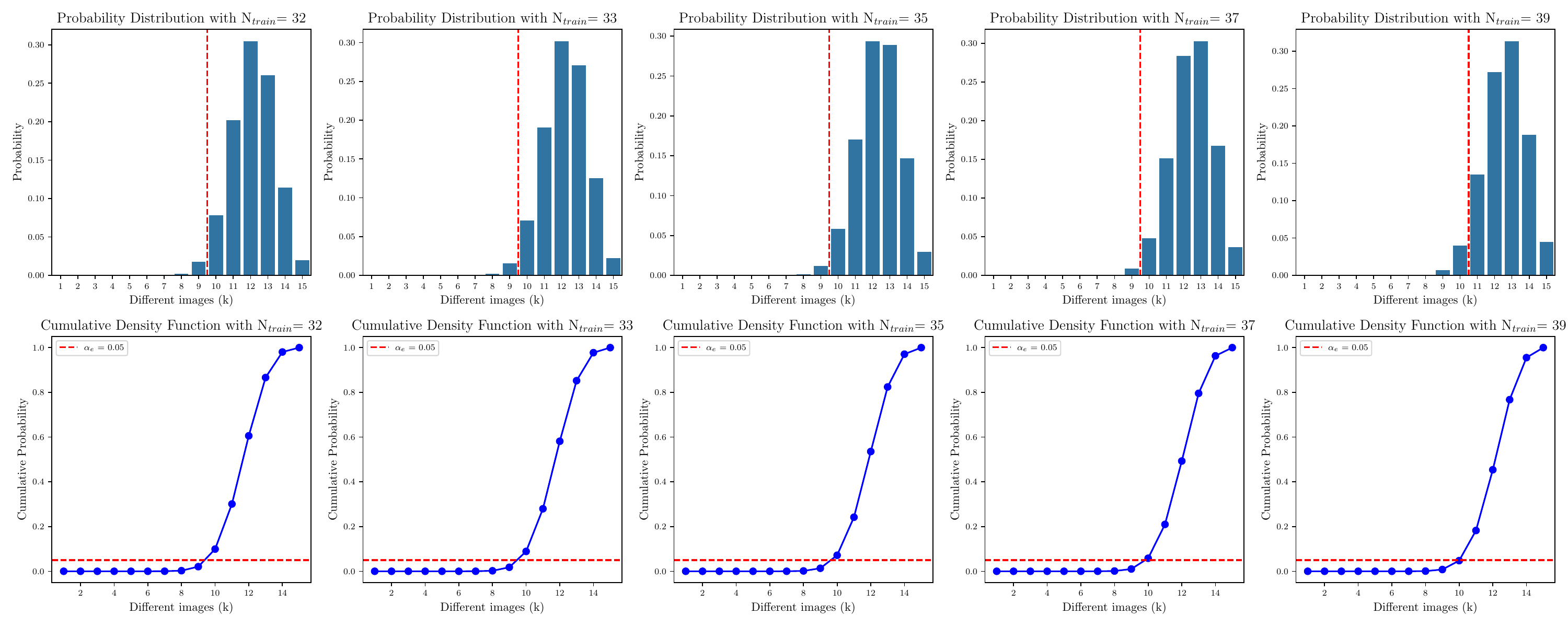}
    \caption{Illustration of the selection process of the threshold for a fixed number of samples with different trainingset sizes.}
    \label{fig:differentsizestoyexample}
\end{figure*}

Next we illustrate the probability distribution function and cumulative density function for five different training sizes in~\cref{fig:differentsizestoyexample}. 
The example uses the real formula for Stirling's number of the second kind instead of the estimate introduced in~\cref{eq:symptotic_est_stir} in the main paper. 
For $\Ntrain = 32$ it is expected that the amount of different images that were sampled after 15 samples lies between 10 and 14. 
If we observe more duplicates than this we would reject the hypothesis that this model has the potential to generate $\Ntrain = 32$ different images. 
For larger training dataset sizes the number of expected different generated images increases. \emph{E.g.}, for $\Ntrain = 39$ we would even reject a model that only produces 10 different images.

\section{Further Visual Results}
\label{sec:visual_results}
\begin{figure*}
    \centering
    \includegraphics[width=0.8\textwidth]{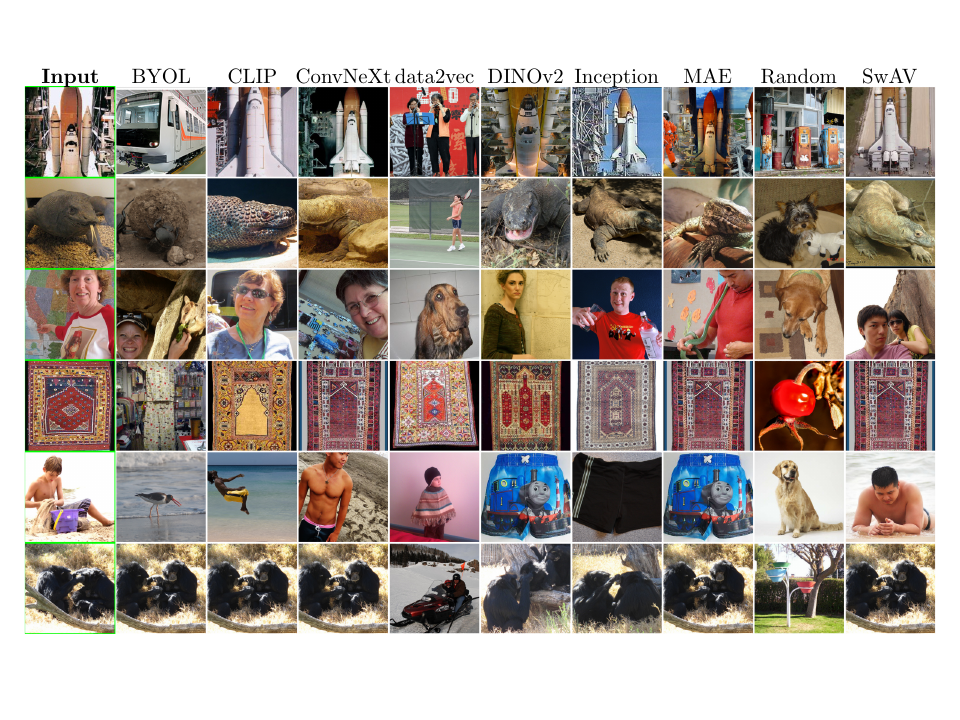}
    \includegraphics[width=0.8\textwidth]{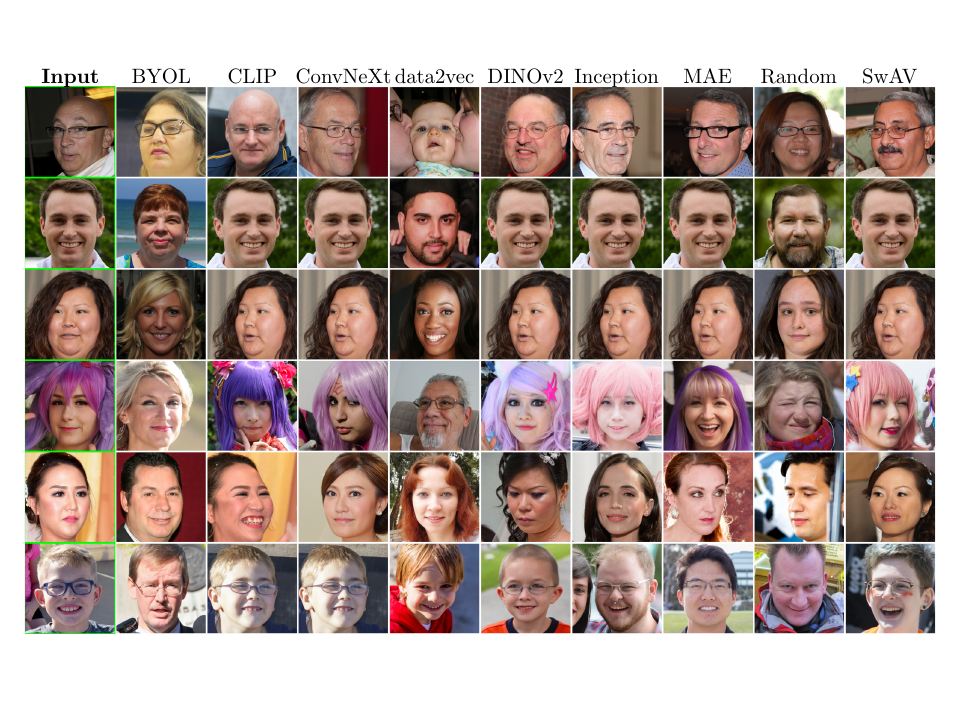}
    \caption{Qualitative results of image retrieval on ImageNet (Top) and FFHQ (bottom).}
    \label{fig:qual_closest_neighbours}
\end{figure*}

To illustrate the image retrieval results, we present visualizations for ImageNet-512 in~\cref{fig:qual_closest_neighbours}. 
For a randomly selected subset of images, we compute the image correspondences based on the predictions $\Pred$ from all benchmark models discussed in~\cref{sec:experiments}.
The results highlight how certain models, such as DINOv2 and Inception, retrieve images based on semantic similarity, while others, like the Random model, focus on general image composition (\emph{e.g.}, a dog resembling a triangular shape against a white background, similar to a child on the beach).
The final example demonstrates that most models retrieve near duplicates, but not all do so consistently. 
DINOv2, for instance, incorporates a deduplication step in its pipeline to minimize redundancy \citep{oquab_dinov2_2024}, which may influence these results.
The third row in~\cref{fig:qual_closest_neighbours} presents an ambiguous case. 
A model that emphasizes image composition, such as DINOv2, may retrieve an image of a woman standing in front of a board. 
In contrast, a model that prioritizes visual similarity in human appearance, such as CLIP, retrieves another image featuring a similarly appearing woman.
This trade-off has to be considered when using feature extractors to assess the performance of generative models with metrics such as FID, Precision, Recall, or IRS.

\section{Image Retrieval Agreement and Consensus}
\label{sec:imageretrieval_agreement_and_consensus}
\begin{figure*}[htbp]
    \centering
    % Subfigure for the Agreement Matrix
     \begin{subfigure}[b]{0.58\textwidth}
        \centering
        \includegraphics[width=\textwidth]{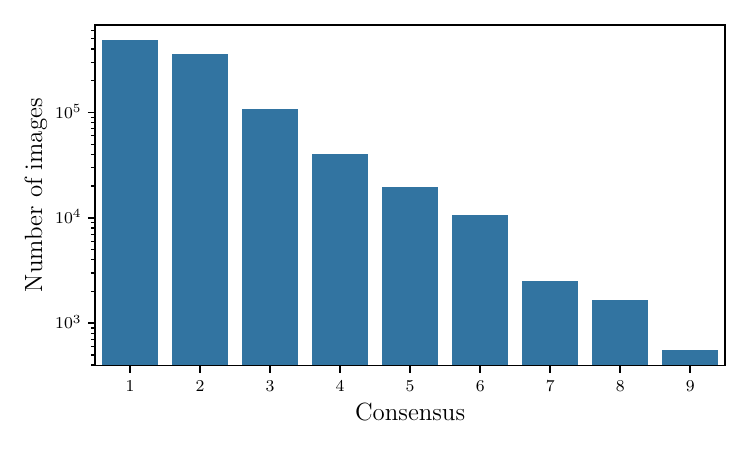}
        %\caption{Consensus for ImageNet}
        \label{fig:imgnet512_consensus_matrix}
    \end{subfigure}
    \hfill
    \begin{subfigure}[b]{0.38\textwidth}
        \centering
        \includegraphics[width=\textwidth]{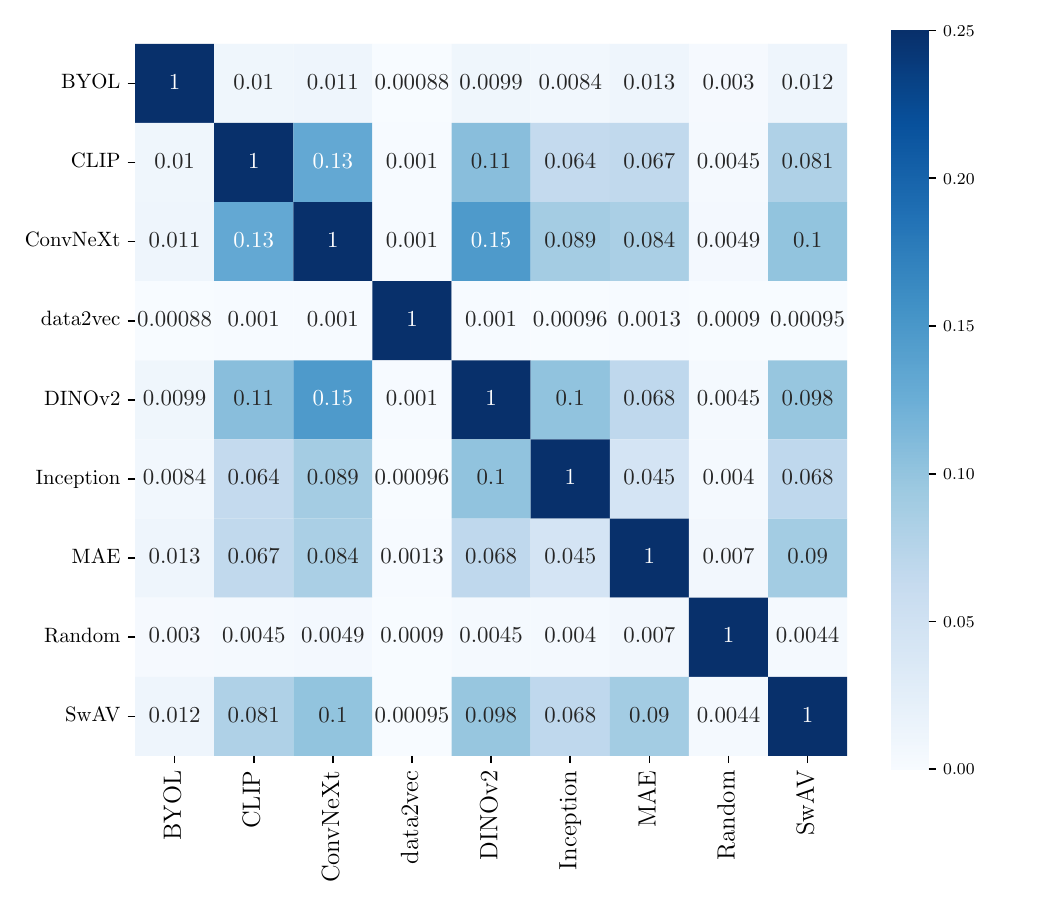}
        %\caption{Agreement Matrix for ImageNet}
        \label{fig:imgnet512_agreement_matrix}
    \end{subfigure}
    \caption{Comparison of Consensus and Agreement of different feature extractors for the ImageNet Dataset}
    \label{fig:imgnet_agreement_and_consensus}
\end{figure*}

In \cref{sec:measurement_gap} we argued that due to the adjustment step introduced in \cref{sec:adjustmentstep} we can in theory choose any kind of feature extractor $\Feat$. 
In order to maximize the interpretability of IRS we use the extractor that has the best agreement with the consensus of all models. 
Therefore, we compute the correspondence prediction for each of the feature extractors for ImageNet. 
We ensemble the decision of each model and call every prediction where five or more models decide for the same image correspondance as consensus. 
Then we check how often each model agrees with the ensemble decision.
Agreement computes how often two models agree with each other. 
Our goal is to see if we can measure how good a model's prediction aligns with the prediction of the ensemble.
The consensus reached by the ensemble is then considered the ground truth. 
The results are shown in \cref{tab:agreemen_with_consensus}. 
There is a large discrepancy between the agreement of all of these models.
Our expectation was that consesus is reached mostly by the same models.
While this is true, the models that are most frequently part of the consensus, are not the models that showed the most diversity which we discuss in \cref{sec:supp_irs_results}. 
DINOv2, for example, performed best in terms of diversity on ImageNet but got beaten on agreement by ConvNeXt, one of the worst models in terms of diversity. 
SwAV on the other hand shows extraordinary agreement with the consensus and is almost always agrees with the consensus, but the output features lack diversity.
Next, we examine ImageNet and the number of feature extractors that agree on image correspondence, as shown in~\cref{fig:imgnet_agreement_and_consensus}. 
We find that, for the majority of images, almost 40\%, all models disagree with each other. 
This outcome aligns with expectations, as many images have multiple valid correspondences, as discussed in~\cref{sec:visual_results}.
The more models required to reach a consensus, the fewer samples meet this criterion. 
Consequently, we consider the consensus decision correct when at least five models agree, as this represents a majority decision.
\cref{fig:imgnet_agreement_and_consensus} further illustrates the level of agreement between models. 
Notably, data2vec and Random exhibit low agreement with all other models, whereas ConvNeXt shows the highest agreement with other models, particularly with CLIP and DINOv2.
This shows that there is no real correlation between diversity and agreement and both have to be measured seperately.

\section{Further Distance Metric Sensitivty Analysis}
We consider two different measurements $\Pred$ to compute the distance between $\fx= \Feat(\xtraininstance)$ of the query image and all reference images  $\fxd = \Feat(\xsynthinstance)$. 
The first one is the cosine distance derived from cosine similarity. 
It is used by many feature extractors directly such as \cite{oquab_dinov2_2024}.
Additionally, we consider the Euclidean distance between features which is used by several metrics such as Precision, Recall \cite{kynkaanniemi_improved_2019}:
\begin{equation}
\begin{split}
 \Pred_{\text{Cosine}}(\xtraininstance, \xsynthinstance) & = d_{\text{Cosine}}(f_{\xtraininstance}, f_{x'}) \\ 
& = 1 - \frac{f_{\xtraininstance} \cdot f_{x'}}{\|f_{\xtraininstance}\| \|f_{x'}\|}
 \label{eq:cosine}
 \end{split}
\end{equation}
and: 
\begin{equation}
\begin{split}
\Pred_{\text{Euclidean}}(\xtraininstance, \xsynthinstance) & =  d_{\text{Euclidean}}(f_x, f_{x'}) \\ 
& = \sqrt{\sum_{i=1}^{n} (f_x^i - f_{x'}^i)^2}
 \label{eq:eucledian}
  \end{split}
\end{equation}
We compute the distances on ImageNet over five folds with a fixed number ratio between the size of training and sampling set.  
The results are shown in \cref{tab:cosinevseucl} .
Generally, we observe the same measurement gap. 
Irrespective of the measurement, all the feature extractors need much longer to converge to 100\%  expected diversity according to \cref{eq:expecteddiversity}.
Visual inspection of the retrieved images from randomly drawn samples yields that the results for each model are very similar. 
However, the performance of some of the models depends heavily on the used metric. 
data2vec, for example, did not show good performance for the Eucledian distance but the best performance for cosine distance. 
DINOv2 is the complete opposite. 
It outperformed all other models according to the Euclidean distance but is only mediocre according to cosine distance.
We conclude that the choice of distance metric has a large influence on the relative performance between all feature extractors.
It does not impact the underlying problem that all of them are far from reaching the expected performance.
We decide to focus our experiments on the Eucledian distance due to its connection to established generative metrics \cite{kynkaanniemi_improved_2019}.

\begin{table}[h!]
\centering
\resizebox{\linewidth}{!}{ % Resize the table to fit the text width
\begin{tabular}{lcccc}
    \toprule
     & \multicolumn{2}{c}{$\alpha = 2$} &  \multicolumn{2}{c}{$\alpha = 6$} \\

    \cmidrule(r){2-3} \cmidrule(l){4-5}
    Idealized                 & \multicolumn{2}{c}{86.47}        & \multicolumn{2}{c}{99.75} \\
      &  $d_{\text{cosine}}$ & $d_{\text{euclidean}}$ &$d_{\text{cosine}}$ &$d_{\text{euclidean}}$ \\
    \cmidrule(r){2-3} \cmidrule(l){4-5}
     
    BYOL      & 67.46 $\pm$ 0.04     &    \textbf{67.93} $\pm$ 0.04     &    87.94 $\pm$ 0.04 & \textbf{88.29} $\pm$ 0.09   \\
    CLIP      & 65.00 $\pm$ 0.03     &    \textbf{66.85} $\pm$ 0.03     &    85.07 $\pm$ 0.05 & \textbf{87.45} $\pm$ 0.08   \\
    ConvNeXt  & \textbf{64.56} $\pm$ 0.03     &    52.34 $\pm$ 0.07     &    \textbf{85.13} $\pm$ 0.09 & 68.30 $\pm$ 0.09   \\
    data2vec  & \textbf{73.08} $\pm$ 0.09     &    40.75 $\pm$ 0.04     &    \textbf{93.26} $\pm$ 0.06 & 59.15 $\pm$ 0.07   \\
    DINOv2    & 66.57 $\pm$ 0.05     &    \textbf{66.64} $\pm$ 0.05     &    \textbf{87.31} $\pm$ 0.06 & \textbf{87.32} $\pm$ 0.04   \\
    Inception & \textbf{65.75} $\pm$ 0.08     &    60.00 $\pm$ 0.05     &    \textbf{86.48 $\pm$} 0.04 & 79.71 $\pm$ 0.10   \\
    MAE       & 68.06 $\pm$ 0.04     &    \textbf{68.17} $\pm$ 0.06     &    88.13 $\pm$ 0.03 & \textbf{88.22} $\pm$ 0.03   \\
    Random    & 69.65 $\pm$ 0.07     &    \textbf{71.29} $\pm$ 0.05     &    89.60 $\pm$ 0.06 & \textbf{90.89} $\pm$ 0.06   \\
    SwAV      & \textbf{70.31} $\pm$ 0.06     &    64.19 $\pm$ 0.08     &    \textbf{90.55} $\pm$ 0.03 & 84.38 $\pm$ 0.16   \\
    
    \bottomrule      
\end{tabular}
}
\label{tab:cosinevseucl}
\caption{Comparison of cosine and Euclidean distances averaged across five folds. 
For each model and $\alpha$, the best $\Pred$ value is highlighted in bold.}
\end{table}

\section{Computational Requirement}
To benchmark the proposed lacking diversity rejection method we use the method for the official ImageNet-512 train set with $\Ntrain = 1281166$. 
We set the desired IRS to 80\% and the probability of error to 5\% with 50000 reference and synthetic samples each. 
\cref{eq:kmin} states that we reject the checkpoint for not being diverse enough if, after sampling, $\Nlearned < 48744$ images are learned (does not account for measurement gap). 
Computing this threshold takes roughly three seconds and does not depend on the images sampled. 
Computing $\Xlearned$ takes longer and depends on the forward pass of $\Feat$.
For Inception-v3 it roughly takes five minutes on a single Nvidia-A40 GPU. 
However, these features are also necessary to compute FID and other metrics so they are usually already available. 
If the pre-computed features are available, computing IRS$_{\infty,a}$ takes 40 seconds for ImageNet. 
Note that we do not need the entire synthetic dataset to compute an upper bound for $\Nlearned$ according to~\cref{eq:Xlearned}. 
If within the first 2000 samples we already observe over 1257 duplicates we can immmediately reject the model. 
For the smaller datasets like Dynamic, computing IRS$_{\infty,a}$ from precomputed features takes less than a second.

\section{Results with Domain Specific Feature Extractors}
\label{sec:supp_results_on_echonet}
In the next step, we analyze the impact of feature extractors on prediction performance.
Specifically, we examine how the performance associated with the observed measurement gap changes when feature extractors are tailored to the dataset.
For this analysis, we compare the privacy models proposed in~\cite{reynaud_echonet-synthetic_2024,dombrowski_uncovering_2024}.
These models were trained for re-identification on the EchoNet dataset \cite{ouyang_video-based_2020}, utilizing a Siamese architecture where the input consists of two frames and the output predicts whether the frames originate from the same video~\cite{dombrowski_uncovering_2024}.
This model can also be directly applied for image retrieval.
The results, presented in \cref{tab:irs-alpha_results_for_echonetdynamic}, show that training models specifically for this dataset reduces the measurement gap.
The models outperform all pre-trained feature extractors by more than eight percentage points.
Furthermore, the findings suggest that the measurement gap can be minimized for specific datasets when necessary.

\begin{table}[]
    \centering
    \resizebox{\linewidth}{!}{
    \begin{tabular}{l|c|c}
        %& $|N_{train}|/ |N_{test}| = 2$ & $|N_{train}|/ |N_{test}| = 16$\\
        $\alpha$ & 2 & 16 \\
        \midrule
        Idealized & 86.47 & 99.99 \\
        \midrule
        BYOL &  69.91 $\pm$ 0.56 & 97.97 $\pm$ 0.48   \\
        CLIP &  64.71 $\pm$ 0.50 & 95.61 $\pm$ 0.74  \\
        ConvNeXt &  56.70 $\pm$ 1.05 & 89.58 $\pm$ 1.12 \\
        data2vec &  56.54 $\pm$ 0.48 & 97.08 $\pm$ 0.62  \\
        DINOv2 &  63.13 $\pm$ 0.33 & 95.15 $\pm$ 0.63 \\
        Inception &  60.62 $\pm$ 0.69 & 93.00 $\pm$ 1.04 \\
        MAE &  64.57 $\pm$ 0.68 & 96.31 $\pm$ 0.81 \\
        Random &  73.35 $\pm$ 0.71 & 98.75 $\pm$ 0.33 \\
        SwAV &  59.20 $\pm$ 0.57 & 92.63 $\pm$ 0.88 \\
        Re-identification \cite{reynaud_echonet-synthetic_2024} & 80.96 $\pm$ 0.34 & 99.61 $\pm$ 0.27 \\
        Re-identification Latent \cite{dombrowski_uncovering_2024} &  \textbf{81.91} $\pm$ \textbf{0.75} & \textbf{99.95} $\pm$ \textbf{0.08} \\
        \bottomrule
    \end{tabular}%
    }
    \caption{IRS$_{\alpha}$ for EchoNet-Dynamic using domain specific re-identification models from \cite{reynaud_echonet-synthetic_2024} and \cite{dombrowski_uncovering_2024}.}
    \label{tab:irs-alpha_results_for_echonetdynamic}
\end{table}

\section{IRS$_\text{real}$ Results}
\label{sec:supp_irs_results}
\begin{figure}
    \centering
    \includegraphics[width=\linewidth]{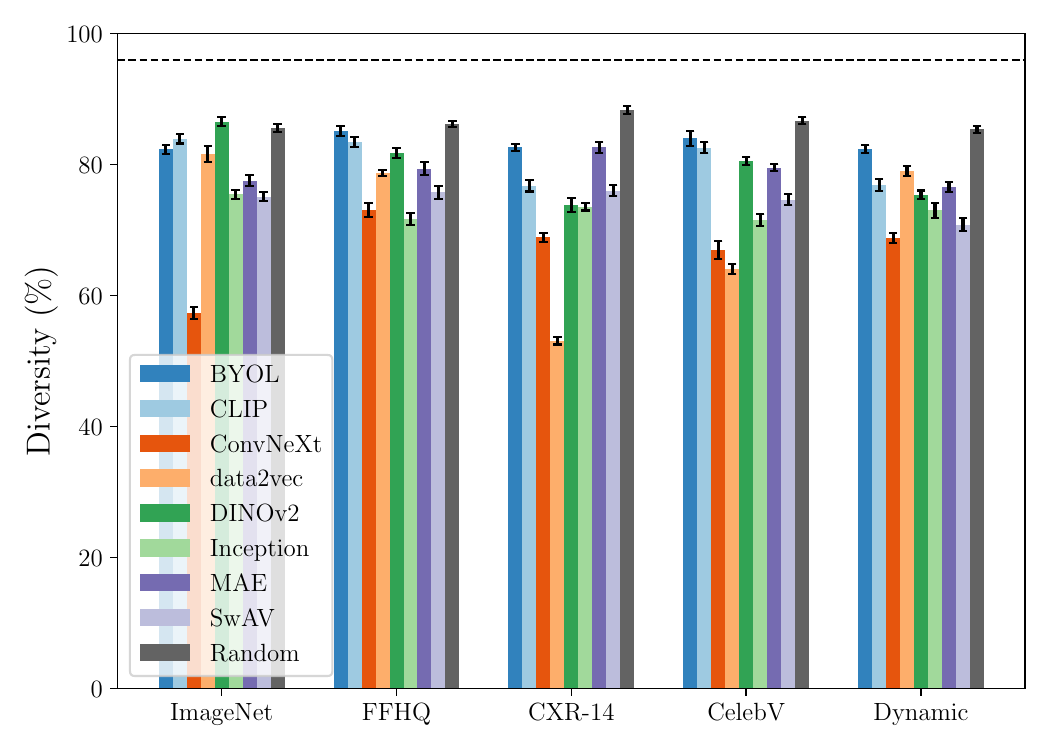}
    \caption{Measured diversity according to IRS of common feature extractors between real training and real test images. The dashed line indicates the idealized model IRS$_\frac{7}{3} = 96.98\%$.}
    \label{fig:barplot-diversity}
\end{figure}

In \cref{sec:measurement_gap} we explain the measurement gap stemming from feature extractors collapsing to smaller features spaces that lack expressiveness in terms of diversity. 
Quantitative prove for this is shown in \cref{fig:barplot-diversity} and \cref{tab:irs}. 
The best performing model across all datasets is the randomly initialized Inception-v3 suggested by \cite{oareilly_pre-trained_2021}.
If we compare this to the visual examples shown in \cref{fig:qual_closest_neighbours}, we see that this similarity seems to be more based on the general composure of the images than the semantics.  
BYOL pre-trained on ImageNet also performs well on all datasets. 
DINOv2 is the best performing model on ImageNet, which confirms the observations from \cite{stein_exposing_2023}.

%\TODO{Short overview of all feature extractors, their training method and why the performance might be good/bad}

\begin{table*}[]
\centering
\begin{tabular}{l|c|c|c|c|c}
 Model & ImageNet & FFHQ & ChestX-ray14 & CelebV-HQ & Dynamic \\
 \toprule
 BYOL      & 82.30 $\pm$ 0.66\%             & \underline{85.13 $\pm$ 0.75\%} & \underline{82.60 $\pm$ 0.53\%}  & \underline{83.96 $\pm$ 1.10\%} & \underline{82.35 $\pm$ 0.59\%}  \\
 CLIP      & 83.92 $\pm$ 0.72\%             & 83.43 $\pm$ 0.74\%             & 76.72 $\pm$ 0.85\%              & 82.54 $\pm$ 0.85\% &                     76.90 $\pm$ 0.92\%      \\
 ConvNeXt  & 57.29 $\pm$ 0.93\%             & 73.04 $\pm$ 1.05\%             & 68.88 $\pm$ 0.70\%              & 66.95 $\pm$ 1.35\% &                     68.82 $\pm$ 0.76\%      \\
 data2vec  & 81.61 $\pm$ 1.23\%             & 78.67 $\pm$ 0.41\%             & 53.07 $\pm$ 0.56\%              & 64.05 $\pm$ 0.76\% &                     78.98 $\pm$ 0.79\%      \\
 DINOv2    & \textbf{86.50 $\pm$ 0.67\%}    & 81.75 $\pm$ 0.70\%             & 73.85 $\pm$ 1.08\%              & 80.53 $\pm$ 0.67\% &                     75.35 $\pm$ 0.67\%      \\
 Inception & 75.46 $\pm$ 0.67\%             & 71.65 $\pm$ 0.92\%             & 73.54 $\pm$ 0.57\%              & 71.53 $\pm$ 0.87\% &                     72.99 $\pm$ 1.14\%      \\
 MAE       & 77.53 $\pm$ 0.80\%             & 79.37 $\pm$ 0.96\%             & \underline{82.62 $\pm$ 0.84\%}  & 79.53 $\pm$ 0.47\% &                     76.58 $\pm$ 0.78\%      \\
 SwAV      & 75.10 $\pm$ 0.68\%             & 75.77 $\pm$ 1.00\%             & 76.00 $\pm$ 0.81\%              & 74.58 $\pm$ 0.84\% &                     70.82 $\pm$ 1.00\%      \\
 Random    & \underline{85.59 $\pm$ 0.63\%} & \textbf{86.16 $\pm$ 0.51\%}    & \textbf{88.32 $\pm$ 0.65\%}     & \textbf{86.67 $\pm$ 0.50\%} &    \textbf{85.38 $\pm$ 0.54\%}     \\
 \bottomrule                                                                 
\end{tabular}                                                     
\caption{IRS computation only using \emph{real} data. 
Results give percentage of real samples retrieved using common feature extractors. 
The idealized scenario reaches IRS$_\frac{7}{3} = 96.98\%$. 
To make comparison across datasets easier results are presented for a fixed size of 3000 training images and 7000 test images on all dataset. 
Best results are bold and second best results underlined. 
}
\label{tab:irs}
\end{table*}

\section{Metrics Analysis and Comparison}
\subsection{
IRS over FID in Measuring Diversity Insufficiency and Bias Amplification}
\label{sec:irs over fid}

Here, we analyze the properties of IRS and demonstrate its superiority over FID~\cite{heusel2017gans} in detecting diversity insufficiency and bias amplification in generative models.

\textit{Definition of FID.} Follow~\cite{heusel2017gans}, here let $\mathcal{X}_r$ and $\mathcal{X}_g$ denote the feature distributions of real and generated data, respectively. The FID is defined as:
\begin{equation}
\text{FID} = \|\boldsymbol{\mu}_r - \boldsymbol{\mu}_g\|^2 + \text{Tr}\left(\mathbf{C}_r + \mathbf{C}_g - 2\left(\mathbf{C}_r \mathbf{C}_g\right)^{1/2}\right)
\end{equation}
where $\boldsymbol{\mu}_r$, $\mathbf{C}_r$ and $\boldsymbol{\mu}_g$, $\mathbf{C}_g$ are the mean vectors and covariance matrices of the real and generated feature distributions, respectively. FID measures the distance between these two distributions under the assumption that they are Gaussian.

\textit{Definition of IRS.} Let $\xtraininstance$ be an image from a dataset consisting of $\Ntrain$ real images residing in image space $\Xtrain \in \mathbb{R}^{c \times h \times w}$. Unconditional generative models aim to learn the distribution $p_{data}(\mathbf{X})$ and sample $\Nsample$ synthetic images from it. Define $\Nlearned$ as the number of unique training samples retrieved by the generated samples. The IRS is:
\begin{equation}
\text{IRS}_\alpha = \frac{\Nlearned}{\Ntrain}
\end{equation}
where $\alpha = \frac{\Nsample}{\Ntrain}$ is the sampling rate. More details can be found in Eq.~\ref{eq:Xlearned}.

\begin{thm} \label{thm1}
IRS exhibits higher statistical sensitivity than FID in detecting diversity insufficiency and bias amplification in generative models.
\end{thm}

\textbf{Proof.} \textit{Diversity Insufficiency.}
Assume the generative model can produce $K < \Ntrain$ unique training samples. The expectation of IRS is:
\begin{equation}
\mathbb{E}[\text{IRS}_\alpha] = \frac{K}{\Ntrain} \left(1 - \left(1 - \frac{1}{K}\right)^{\Nsample}\right)
\end{equation}
Since $K < \Ntrain$ and $\left(1 - \frac{1}{K}\right)^{\Nsample} > \left(1 - \frac{1}{\Ntrain}\right)^{\Nsample} \approx e^{-\alpha}$, it follows that:
\begin{equation}
\mathbb{E}[\text{IRS}_\alpha] < 1 - e^{-\alpha}
\end{equation}
Thus, IRS effectively captures the reduction in diversity when $K < \Ntrain$.

Conversely, FID measures the distance between feature distributions based on mean and covariance. Even if $K < \Ntrain$, if the $K$ samples are diverse in feature space, $\boldsymbol{\mu}_g$ and $\mathbf{C}_g$ may remain close to $\boldsymbol{\mu}_r$ and $\mathbf{C}_r$, resulting in a low FID that fails to reflect the reduced diversity.

\textit{Bias Amplification.} Suppose the generative model memorizes (\emph{i.e.}, guided) $K \ll \Ntrain$ training samples, effectively reproducing these samples. The expectation of IRS is:
\begin{equation}
\mathbb{E}[\text{IRS}_\alpha] = \frac{K}{\Ntrain} \left(1 - \left(1 - \frac{1}{K}\right)^{\Nsample}\right) \ll 1
\end{equation}
Given $K \ll \Ntrain$, IRS approaches $\frac{K}{\Ntrain}$, significantly lower than the ideal value of $1 - e^{-\alpha}$, thereby effectively indicating bias amplification.

In contrast, 
$
\text{FID} \approx \|\boldsymbol{\mu}_r - \boldsymbol{\mu}_g\|^2 + \text{Tr}\left(\mathbf{C}_r + \mathbf{C}_g - 2\left(\mathbf{C}_r \mathbf{C}_g\right)^{1/2}\right).
$
If the memorized $K$ samples have feature statistics close to the biased data distribution, FID remains low, failing to detect bias amplification.

\subsection{Comparison to Other Metrics}
\label{sec:other_metrics_hyperparams}
\begin{figure}
    \centering
    \includegraphics[width=0.85\linewidth]{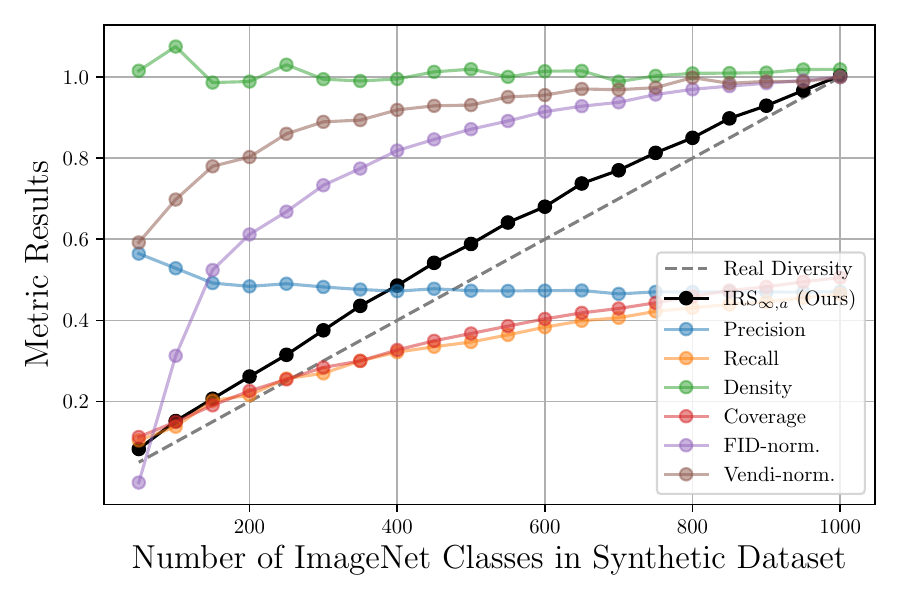}
    \caption{Measuring diversity of datasets by removing classes and computing IRS.}
    \label{fig:metric_interpretable_k1}
\end{figure}

Alternative metrics, such as Precision, Recall, Density, and Coverage, are susceptible to issues arising from varying hyperparameter settings. 
We demonstrate this by setting the number of considered neighbors to 1 for these metrics and replicating the experiment shown in \cref{fig:metric_interpretable}. 
The results, presented in \cref{fig:metric_interpretable_k1}, reveal that all metrics converge to very low diversity values, even when only real data is used. This indicates that incorrect hyperparameter configurations can prevent these metrics from converging to a diversity value of 1.0 for real data, significantly compromising the interpretability of the results.

\end{document}